%% file: ms.tex
\documentclass[10pt,twocolumn,letterpaper]{article}
\usepackage[english]{babel}
\usepackage[utf8x]{inputenc}
\usepackage[T1]{fontenc}
\usepackage{balance}
\usepackage{times}
\usepackage{cvpr}
\usepackage{epsfig}
\usepackage[pagebackref=true,breaklinks=true,letterpaper=true,colorlinks,bookmarks=false]{hyperref}
\usepackage[leqno]{amsmath}
\usepackage{amssymb}
\usepackage{subcaption} 
\usepackage{graphicx}
\usepackage{multirow}
\usepackage[colorinlistoftodos]{todonotes}
\usepackage{wrapfig}
\usepackage{pgfplots} 
\usepackage{tikz} 
\usepackage{enumitem}
\usetikzlibrary{shapes, shapes.geometric, positioning, calc, intersections, patterns, pgfplots.groupplots}
\usepackage{tkz-euclide}
\usetkzobj{all}

\usepackage{math-style}

\newcolumntype{C}[1]{>{\centering\arraybackslash}p{#1}}

\cvprfinalcopy %*** Uncomment this line for the final submission

\def\cvprPaperID{3631} % *** Enter the CVPR Paper ID here

\ifcvprfinal\fi

\makeatletter
\newcommand{\leqnomode}{\tagsleft@true}
\newcommand{\reqnomode}{\tagsleft@false}
\makeatother

\def\oover{\abovewithdelims...1pt}

\makeatletter
\newcommand\footnoteref[1]{\protected@xdef\@thefnmark{\ref{#1}}\@footnotemark}
\makeatother

\newcommand{\ARXIV}[2]{#1} % for ARXIV

\makeatother
\begin{document}

%\title{Superpoint Oversegmentation with Graph-Structured Deep Metric Learning}
%\title{3D-Superpoint Oversegmentation with Graph-Structured Deep Metric Learning}
\title{Point Cloud Oversegmentation with Graph-Structured Deep Metric Learning}
\author{Loic Landrieu$^{1}$, Mohamed Boussaha$^{2}$\\
Univ. Paris-Est, IGN-ENSG, LaSTIG, \textsuperscript{1}STRUDEL, 
\textsuperscript{2}ACTE, Saint-Mand\'e, France\\
{\tt\small loic.landrieu@ign.fr,mohamed.boussaha@ign.fr}}

\maketitle

\begin{abstract}We propose a new supervized learning framework for oversegmenting 3D point clouds into superpoints. We cast this problem as learning deep embeddings of the local geometry and radiometry of 3D points, such that the border of objects presents high contrasts. The embeddings are computed using a lightweight neural network operating on the points' local neighborhood. Finally, we formulate point cloud oversegmentation as a graph partition problem with respect to the learned embeddings.

This new approach allows us to set a new state-of-the-art in point cloud oversegmentation by a significant margin, on a dense indoor dataset (S3DIS) and a sparse outdoor one (vKITTI). Our best solution requires over five times fewer superpoints to reach similar performance than previously published methods on S3DIS. Furthermore, we show that our framework can be used to improve superpoint-based semantic segmentation algorithms, setting a new state-of-the-art for this task as well.
\end{abstract}
%==========================================================================
\section {Introduction}
%==========================================================================
%
The interest of segmenting point clouds into sets of points known as superpoints---the 3D equivalent of superpixels--- as a preprocessing step to their analysis has been extensively demonstrated \cite{landrieu2017large, rusu2008towards,pu2006automatic,chen2008architectural,xiong20113}.
However, these unsupervized methods rely on the assumption that segments which are geometrically and/or radiometrically homogeneous are also semantically homogeneous. This assertion should be challenged, especially since the quality of any further analysis is limited by the quality of the initial oversegmentation. Our objective in this paper is to formulate a supervized framework for oversegmentating 3D point clouds into semantically pure superpoints in order to facilitate their semantic segmentation.

Although superpixel-based methods and deep learning have both been around for a long time in computer vision, convolutional neural networks have only recently been used for superpixel oversegmentation. Notably, \cite{liu2018learning} introduced a loss function emulating oversegmentation metrics, and which is compatible with graph-based clustering methods. \cite{JampaniSLYK18} propose a fully differentiable version of the SLIC superpixel algorithm \cite{achanta2012slic}, allowing for end-to-end training of spatial clustering methods. Both approaches have shown promising results, displaying significant improvement upon methods relying on handcrafted descriptors. In this paper, we build upon these ideas, albeit in the 3D setting.

We propose formulating point cloud oversegmentation as a deep metric learning problem structured by an adjacency graph defined on an input 3D point cloud.
We introduce the \emph{graph-structured contrastive loss}, a loss function which learns to embed 3D points homogeneously within objects and with high contrast at their interface. This loss can be adapted to the non-differentiable task of oversegmentation by using our \emph{cross-partition weighting} strategy. The points' embeddings themselves are computed from the points' local geometry and radiometry by a lightweight model inspired from PointNet \cite{qi2017pointnet} and called \emph{Local Point Embedder} ($\LCE$). 
Finally, the superpoints are defined as a piecewise-constant approximation of the learned embedding in the adjacency graph, in the manner of \cite{guinard2017weakly}.
\setlength\tabcolsep{0pt}
\begin{figure*}[!ht]\centering
\begin{tabular}{cccc}
\begin{subfigure}[b]{0.245\textwidth}
\includegraphics[width=1\textwidth]{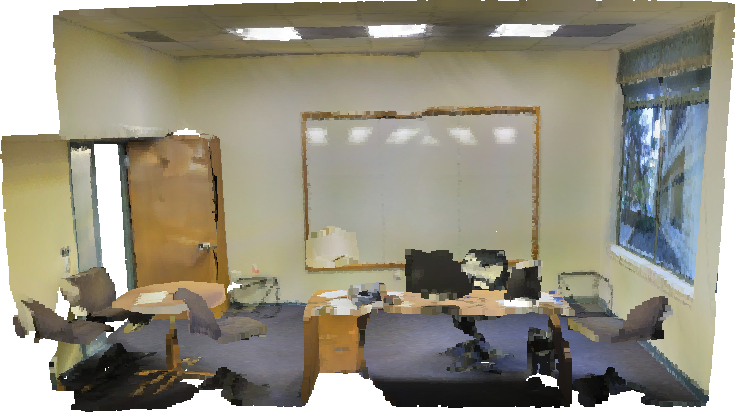}
\caption{Input Point Cloud}
\label{fig:introfig_rgb}
\end{subfigure}
&
\begin{subfigure}[b]{0.245\textwidth}
\includegraphics[width=1\textwidth]{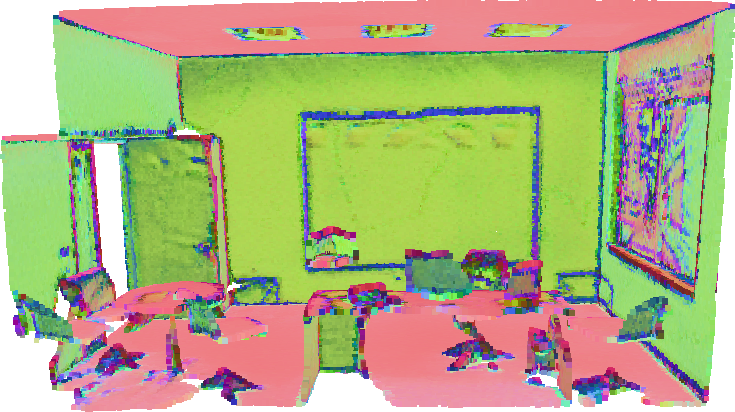}
 \caption{Learned Embedding}
\label{fig:introfig_emb}
\end{subfigure}
&
\begin{subfigure}[b]{0.245\textwidth}
\includegraphics[width=1\textwidth]{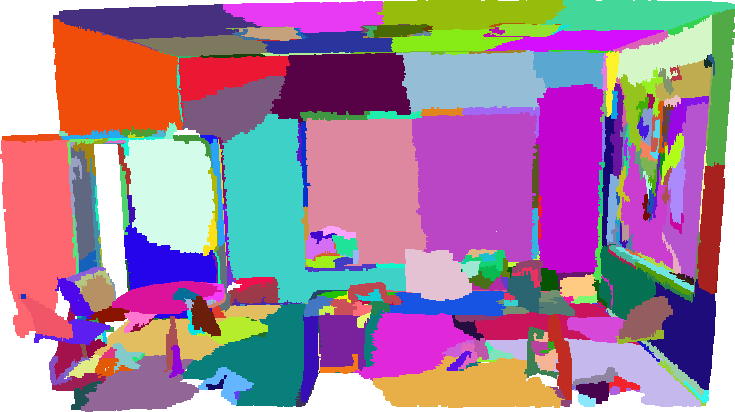}
\caption{Oversegmentation}
\label{fig:introfig_seg}
\end{subfigure}
&
\begin{subfigure}[b]{0.245\textwidth}
\includegraphics[width=1\textwidth]{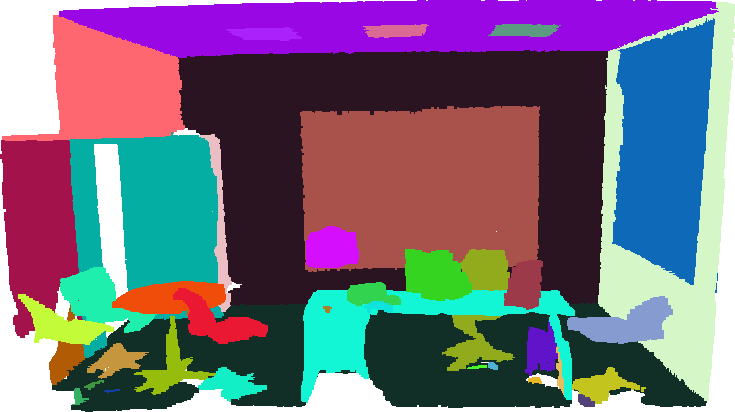}
\caption{True Objects}
\label{fig:introfig_gt}
\end{subfigure}
\end{tabular}
\caption{Illustration of our framework on a hard-to-segment scene with a white board on a white wall: a colored point cloud is given as input \Subref{fig:introfig_rgb}, an embedding is computed for each point \Subref{fig:introfig_emb}, which allows a clustering technique to compute an oversegmentation \Subref{fig:introfig_seg}, which closely follows the ground truth \Subref{fig:introfig_gt}. Throughout the figures of this paper, the embeddings are projected into a 3-dimensional space to allow color visualization.
}
\label{fig:illu_all}
\end{figure*}
%--------------------------------------------------------------------------

Furthermore, we define the end-goal of our point cloud oversegmentation as assisting semantic segmentation methods by providing semantically pure superpoints. 
We show that our approach can be integrated with the superpoint graph approach of \cite{landrieu2017large}  to significantly improve the partition step, and consequently the resulting semantic segmentation. The contributions of this paper are as follows:
\begin{itemize}[topsep=3pt,itemsep=0pt]
\item %To the best of our knowledge,
We present the first supervized framework for 3D point cloud oversegmentation;
\item We introduce the graph-structured contrastive loss, which can be combined with our cross-partition weighting strategy to produce point embeddings with high contrast at objects' borders;
\item We introduce the local point embedder, a lightweight architecture, inspired by \cite{qi2017pointnet}, to embed the local geometry and radiometry of 3D points in a compact way;
\item We significantly improve the state-of-the-art of point cloud oversegmentation for two well-known and very different datasets;
\item When combined with the superpoint graph semantic segmentation method, our approach improves upon the state-of-the-art for this task as well.
\end{itemize}
%--------------------------------------------------------------------------
%==========================================================================
\section {Related work}
%==========================================================================
\textbf{Superpixels/ Supervoxels}: 
There is a large body of literature on the oversegmentation of images into superpixels \cite{StutzHL18} and videos into supervoxels \cite{XuC12}. These methods can be divided into two groups: graph-based, which exploit the pixels' connectivity \cite{FelzenszwalbH04,GrundmannKHE10a,liu2011entropy}, and cluster-based, which use the pixels' relative positions \cite{achanta2012slic,BerghBRG15,YaoBFU15,levinshtein2009turbopixels}. Recently, deep learning methods have been successfully used to develop supervized superpixels oversegmentation approaches, either graph-based \cite{liu2018learning}, or cluster-based \cite{JampaniSLYK18}. 

\textbf{Oversegmentation of 3D Point Clouds}:
The aforementioned methods perform well on images, but rely on the regular structure of pixels. 3D point clouds, as unordered point sets with irregular distributions, require special attention.
\cite{BENSHABAT2018} propose three extensions of 2D local variation graph-based method \cite{FelzenszwalbH04} to 3D oversegmentation and study different strategies for constructing the graph, edge weights, and subgraph merging. \cite{song2014} introduce a graph-structured approach which exploits the structure of LiDAR sensors to remove edges corresponding to boundary points.
\cite{PaponASW13} propose a cluster-based method based on the $k$-means algorithm and octrees. However, this method remains  sensitive to the clusters' initialization.
\cite{GaoLZF17} use the visual saliency of RGBD images to initialize clustering. \cite{LIN201839} propose a clustering method which does not require such initialization, and is therefore less sensitive to the irregular densities of LiDAR point clouds.
Likewise, \cite{guinard2017weakly} introduce an initialization-free segmentation model formulated as a graph-structured optimization problem.
All these methods rely on hand-crafted geometric and/or colorimetric features.

\textbf{Deep Learning for 3D Point Clouds}: The work in \cite{qi2017pointnet} has pioneered the use of deep learning for 3D point cloud processing. However, this usage has so far only been used for semantic segmentation \cite{li2018pointcnn,tchapmi2017segcloud,EngelmannKHL17_vkitty,simonovsky2017dynamic,Riegler2017OctNet,QiYSG17PointNetPP,ye20183d,wang2018deep}, object detection \cite{zhou2017voxelnet}, or reconstruction \cite{groueix2018atlasnet}.
To the best of our knowledge, no supervised 3D point oversegmentation technique that leverages deep learning-based embeddings to generate superpoints has been developed yet.

\textbf{Metric Learning}: 
Metric learning aims to learn a similarity function between data points with properties corresponding to a given task \cite{kulis2013metric}.
In practice, an embedding function associates each data point with a feature vector attuned to a given objective. These objectives can be related to classification \cite{goldberger2005neighbourhood, salakhutdinov2007learning}, or clustering \cite{song2016learnable, hershey2016deep}, among many other applications (see \cite{aljalbout2018clustering} for a useful taxonomy).
In the context of deep learning, this can be achieved by using a well-chosen loss, such as the contrastive loss \cite{chopra2005learning, bromley1994signature}; the triplet loss \cite{hoffer2015deep} or some of its variants \cite{wang2017deep}.
Notably, metric learning has recently been used to improve the quality of learned features for a 3D point semantic segmentation task \cite{engelmann2018}.
However, our task is different in the sense that our embeddings are related to oversegmentation through a graph partition problem rather than classification.
%==========================================================================
\section {Method}
%==========================================================================
 %When considering the task of segmenting a room, this can be understood as first painting every objects with colors such that no adjacent object share the same color, and then using this coloring to infer the limits between objects.
Our goal is to produce a high-quality 3D-point cloud oversegmentation, so that it can be in turn used by superpoint-based semantic segmentation algorithms. This translates into the following three properties:
\begin{enumerate}[parsep=1pt,itemsep=1pt, label=($\cP$\arabic*),leftmargin=*]
\item \label{prop:p1} \textbf{object-purity}: superpoints must not overlap over objects, especially if their semantics are different;
\item \label{prop:p2} \textbf{border recall}: the interface between superpoints must coincide with the borders between objects;
\item \label{prop:p3} \textbf{regularity}: the shape and contours of the superpoints must be simple.
\end{enumerate}
Our approach can be broken down into two steps: in \secref{sec:emb} we present the local cloud embedder, a simple neural network which associates each point with a compact embedding that captures its local geometry and radiometry. In \secref{sec:part}, we describe how we compute a point cloud oversegmentation from this embedding using either graph or cluster-based oversegmentation algorithms.

Throughout this paper we will stress the difference between \emph{set-features}, which are unordered sets of descriptors (such as information related to the neighbors of a point), and \emph{point-features}, which characterize a specific point. Set features will always be capitalized, while point-features will use lowercase.

Let us consider a point cloud $C$, with each point $i$ defined with its position $p_i \in \bbR^3$ and $d$-dimensional radiometric information $r_i \in \bbR^d$ (this can be colors if available, or intensity for LiDAR scans, or be ignored if none is available).
Each point $i$ is associated with the set-features $P_i$ and $R_i$, respectively comprised of the position and radiometry of its $k$ nearest neighbors $N_i$ in the input cloud: $P_i = \Cur{p_j \mid j \in N_i}, R_i= \Cur{r_j \mid j \in N_i}$. 
For ease of notation, any operator or function $f$ applied to a set-feature $X$ is to be understood as being applied to all its elements: $f(X)=\Cur{f(x) \mid x \in X}$. 
%.....................................................................
\subsection  {Local Point Embedding }
%.....................................................................
\label{sec:emb}
%---------------------------------------------------------------------
%\addtolength{\belowcaptionskip}{-1mm}
\addtolength{\abovecaptionskip}{-6.5mm}
\begin{figure}
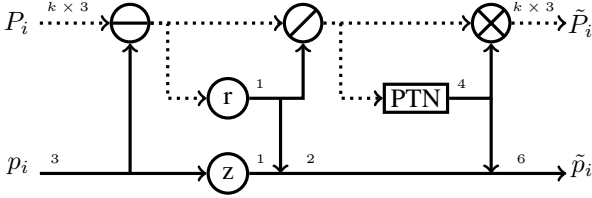

\include{figure_STN}
\caption{Architecture of the spatial transform network. It takes a point's coordinate as point-input $p_i$ and the coordinates of its neighbors as set-input $P_i$. The vertex $r$ computes the radius of a point cloud \eqref{eq:diam}, the vertex $z$ extract the vertical coordinate of a point's position, and the vertex $\PTN$ is a small PointNet-like network \eqref{eq:st} which outputs a $2\times 2$ rotation matrix around the $z$ axis \eqref{eq:rot}. In this and subsequent figures, set-features (respectively point-features) are represented by a dotted line (respectively a solid line). The numbers above the lines represent the size of the channels.}
\label{fig:stn}
\end{figure}
%--------------------------------------------------------------------------
%
%----------------------------------------------------------------------
%----------------------------------------------------------------------
\begin{figure}
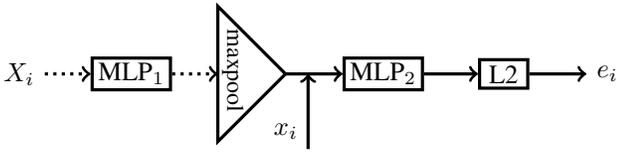

\include{figure_RCE}
\caption{Architecture of the local point embedder ($\LCE$) \eqref{eq:ptn}, which computes an embedding set-feature $X_i$ and point-feature $x_i$ encoding the local radiometry and the normalized geometry. The $\Ltwo$ block normalizes the output on the unit sphere \eqref{eq:ltwo}.}
\label{fig:RCE}
\end{figure}
\addtolength{\abovecaptionskip}{5mm}
%----------------------------------------------------------------------
Our objective is to associate to each point a compact $m$-dimensional embedding $e_i$ characterizing its point-features (position, color, etc.) and the geometry and radiometry of its local neighborhood. The embeddings are constrained to be within the $m$-unit sphere $\bbS_m$, as suggested by \cite{wang2014learning}, to prevent collapse during the training phase, and to normalize their distance with one another.

To this end, we introduce the Local Point Embedder ($\LCE$), a lightweight network inspired by PointNet \cite{qi2017pointnet}. However, unlike PointNet, $\LCE$ does not try to extract information from the whole input point cloud, but rather encodes each point based on purely local information. Here, we describe the different units of our network.\\
\textbf{Spatial Transform:} This unit takes the positions of a target point $p_i$ and its local $k$-neighborhood $P_i$, as represented in \figref{fig:stn}. It normalizes the neighbors' coordinates around $p_i$, and such that the standard deviation of the point's position is equal to $1$ \eqref{eq:norm}. Then, this neighborhood is rotated around the $z$ axis with a $2\times2$ rotation matrix computed by small PointNet network $\PTN$ \eqref{eq:rot}. As advocated by \cite{jaderberg2015spatial}, these steps aim to standardize the position of the neighborhood clouds of each point. This helps the next network to learn position distribution. 
Along the normalized neighborhood position $\Pit$, this unit also outputs geometric point-features $\pit$ describing the elevation $\upp{p}{z}_i$, the neighborhood radius, as well as its original orientation (through the $4$ values of the rotation matrix: $[\Omega_{x,x},\Omega_{x,y},\Omega_{y,x},\Omega_{y,y}]$)\eqref{eq:pout}. By keeping track of the normalization operations, the embedding can stay covariant with the original neighborhood's radius, height, and original orientation, even though the points' positions have been normalized and rotated.\vspace{-5mm}
\vspace{-5mm}
\setlength\tabcolsep{0pt}
\begin{tabular}{p{0.43\columnwidth}p{0.57\columnwidth}}
      {\leqnomode \begin{align}%\nonumber%\label{eq:diam}
              & \rad  =  \std{P_i} \label{eq:diam} \\%\nonumber%\label{eq:stn}
              \label{eq:st}
              & \Omega = \PTN(\Pit) \\\label{eq:norm}
              & P_i'  =  (P_i - p_i) /\rad
               \end{align}}
        & {\reqnomode \begin{align}\label{eq:rot}
               &\Pit  =  \{p \times \Omega \mid p \in P_i'\} \\\label{eq:pout}
               &\pit  =  [p_i^{(z)}, \rad, \Omega]%\\%\label{eq:stn}
               &%\Pit, \pit = ST(P_i, p_i)
               \end{align}}
\end{tabular}
\setlength\tabcolsep{5pt}
\textbf{Local Point Embedder:} 
The $\LCE$ network, represented in \figref{fig:RCE}, computes a normalized embedding from two inputs: a point-feature $x_i$ and a set-feature $X_i$. As in PointNet \cite{qi2017pointnet}, the set-features are first processed independently by a multi-layer perceptron (denoted $\MLP_1$) comprised of a succession of layers in the following order: linear, activation (ReLu \cite{nair2010rectified}), normalization (batch \cite{ioffe2015batch}), and so on. The resulting set-features are then maxpooled into a point-feature, which is concatenated with the input point-feature. The resulting vector is processed through another multi-layer perceptron $\MLP_2$ \eqref{eq:ptn}, and finally normalized on the unit sphere.

%\vspace{-3mm}
The embeddings $e_i$ are computed for each point $i$ of $C$ through a shared $\LCE$ \eqref{eq:emb}. The input set-feature $X_i$ is set as the concatenation of the neighbour's transformed position $\Pit$ and their radiometric information $R_i$, while the input point-feature $x_i$ is composed of the neighborhood geometric point-feature $\pit$ and the radiometry $r_i$ of  point $i$.
\reqnomode
\begin{flalign}\label{eq:ltwo}
&\!\!\!\!\!\!\!\!\!\!\Ltwo(\cdot)=\cdot/\Vert \cdot \Vert\\\label{eq:ptn}
&\!\!\!\!\!\!\!\!\!\!\LCE(X_i,x_i)\!=\!\Ltwo\Pa{\MLP_2\Pa{[\max\Pa{\MLP_1(X_i)}, x_i]}}\\\label{eq:emb}
&\!\!\!\!\!\!\!\!\!\!e_i=\LCE([\Pit,R_i],[\pit,r_i])
\end{flalign}
%----------------------------------------------------------------------
%......................................................................
\subsection  {Graph-Based Point Cloud Oversegmentation}
%......................................................................
\label{sec:part}
%......................................................................
\subsubsection{The Generalized Minimal Partition Problem}
Once the embeddings are computed, we define the superpoints with respect to an adjacency graph $G=(C,E)$ derived from the point cloud $C$. Note that $E$ can be obtained from the neighbors' structure used for the $\LCE$. However, we find that much smaller neighborhoods are needed to capture the cloud's adjacency structure than to describe the local neighborhood of points.
As proposed by \cite{guinard2017weakly},   
we define the superpoints as the constant connected components in $G$ of a piecewise-constant approximation of the embeddings $e\in \bbS_m^C$.
This approximation is the solution $f^\star$ of the following optimization problem:
\reqnomode
\reqnomode\begin{flalign}
\!\!\!\!\!\!\!\!\!\!\!\!\!\!\!f^\star = \argmin_{f \in \bbR^{C \times m}}
\sum_{i\in C}\!\Vert f_i - e_i \Vert^2 +\!\!
\sum_{(i,j)\in E}w_{i,j} \Bra{f_i \neq f_j},
\label{eq:mgp}
\end{flalign}
with $w \in \bbR_+^{E}$ the edges' weight and $\Bra{ x \neq y}$ equal to $0$ if $x=y$ and $1$ otherwise.
To encourage the network to split along high contrast areas, we define the edge weight as
$
w_{i,j} = \lambda \exp \Pa{\frac{-1}{\sigma} \Vert e_i - e_j \Vert^2},
$
with parameters $\lambda,\sigma\in\bbR^+$.

Problem \eqref{eq:mgp}, known as the \emph{generalized minimal partition} (GMP) and introduced by \cite{landrieu2017cut}, is neither continuous, differentiable, nor convex, and therefore the global minimum cannot be realistically retrieved. However, the $\ell_0$-cut pursuit algorithm \cite{landrieu2017cut}
allows for fast approximate solutions.

The contour penalty automatically implements \ref{prop:p3} for reasonable parameterization of the problem.
Note that the optimization variable $f$ can take its values in $\bbR^{C \times m}$, while each embedding $e_i$ is constrained on the $m$-sphere. This is a limitation of our approach due to efficiency concerns. It can lead to some suboptimal approximate solutions. However, we show in the numerical experiments that the learned embeddings lead to satisfactory partitions.%next part how we can mitigate this within the learning process.

%....................................................................
\subsubsection{Graph-Structured Contrastive Loss}
%...................................................................
\label{sec:graph}
%\MO{
%...................................................................
%\label{sec:graph}
%As mentioned earlier, \eqref{eq:mgp} is a non-continuous non-convex optimization problem. Moreover computing connected components on a graph is inherently non-differentiable. That is why, instead of  directly optimizing \eqref{eq:mgp} to ensure the semantic purity property \ref{prop:p1}, since it is the principal quality of compact superpoints, we note that if the border recall property \ref{prop:p2} is implemented (\ie superpoints and objects share the same boundaries), then \ref{prop:p1} ensues. Therefore, we propose a surrogate loss called \emph{graph-structured contrastive loss} focusing on correctly detecting the borders between objects of a point cloud $C$ structured by the adjacency graph $G$. 
%We name $\Eintra$ (resp. $\Einter$ ) the set of \emph{intra-edges}  (resp. \emph{inter-edges}) the subset of edges of $G$ between points within the same object (resp. point from different adjacent objects). 
%}

As mentioned earlier, the semantic purity property \ref{prop:p1} is the first quality of superpoints. Once could imagine taking a metric estimating the semantic purity of the solution of \eqref{eq:mgp} as a loss function. However, the GMP is a non-continuous non-convex optimization problem, and computing connected components on a graph is inherently non-differentiable. This makes optimizing directly with respect to properties of the partition very hard, if not impossible.

Instead, we note that if the border recall property \ref{prop:p2} is implemented (\ie superpoints and objects share the same boundaries), then \ref{prop:p1} ensues. Therefore, we propose a surrogate loss called \emph{graph-structured contrastive loss} focusing on correctly detecting the borders between objects. 
To this end, we define $\Eintra$ (resp. $\Einter$ ) the set of \emph{intra-edges}  (resp. \emph{inter-edges}) as the set of edges of $G$ between points within the same object (resp. point from different adjacent objects).

In the spirit of the original contrastive loss \cite{chopra2005learning}, our loss encourages embeddings of vertices linked by an intra-edge to be similar, while rewarding different embeddings when linked by an inter-edge:
\reqnomode\begin{align}\nonumber
\!\ell(e) \!= \!\frac1{\vert E\vert}
\Pa{
\sum_{(i,j)\in \Eintra}\!\!\!\!\!\phi\Pa{e_i - e_j}+
\!\!\!\!\!\!\sum_{(i,j)\in \Einter}\!\!\!\!\!\! \mu_{i,j} \psi\Pa{e_i - e_j}}\!,
%\label{eq:gtl}
\end{align}
with $\phi$ (resp. $\psi$)  a function minimal (resp. maximal) at $0$, and $\mu_{i,j} \in \bbR^{\Einter}$ a weight on inter-edges. A point embedding function minimizing this loss will be uniform within objects and have stark contrasts at their interface. Consequently, the components of the piece-wise constant approximation of \eqref{eq:mgp} should follow the objects' borders.
%--------------------------------------------------------------------
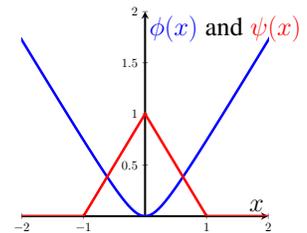
\begin{wrapfigure}{R}{0.23\textwidth}
\resizebox{0.23\textwidth}{!}{
\begin{tikzpicture}
\begin{axis}[ xmin = -2, xmax = 2, ymin = 0, ymax = 2, xlabel=\huge $x$, ylabel={\huge\textcolor{blue}{$\phi(x)$}\;and\;\textcolor{red}{$\psi(x)$}},axis line style = ultra thick,axis lines=middle]
\addplot[blue,line width = 2pt,mark=none,samples=200] {0.3 * ((((x/0.3)^2 + 1)^0.5)-1)};
\addplot[color=red,mark=none,line width = 2pt] coordinates { (-2,0) (-1,0) (0,1) (1,0) (2,0)};
\end{axis}
\end{tikzpicture}
}
\caption{The functions $\phi$ (in blue) and $\psi$ (in red) used in the graph-structured contrastive loss.}
\label{fig:huber}
\end{wrapfigure}
%--------------------------------------------------------------------
This loss differs from the triplet loss \cite{hoffer2015deep,wang2014learning}, as it involves all vertices within a graph (or a sub-graph) at once, and not just an anchor and related positive/negative examples. In this way, it bypasses the problem of example picking altogether. Indeed, the positive and negative examples are directly given by the adjacency structure set by $\Eintra$ and $\Einter$. It differs from \cite{engelmann2018} as it does not try to learn semantic information, but rather to compute a signal on a graph such that its constant approximation respects certain properties, with no attention to semantics. Indeed, objects of different classes can share the same embeddings as long as they are never adjacent, such as floors and ceilings for indoor scenes.

We chose $\phi$, the function promoting intra-object homogeneity as $\phi(x)=\delta (\sqrt{\Vert x \Vert^2 / \delta^2+1}-1)$ with $\delta=0.3$ (represented in \figref{fig:huber}). This means that the first term of $\ell$ is the (pseudo)-Huber graph-total variation on the $\Eintra$ edge \cite{huber1973robust, charbonnier1997deterministic}, promoting smooth homogeneity of embeddings within the same object.

With $\psi(x)=\max\Pa{1-\Vert x \Vert ,0}$, the second part of $\ell$ is the opposite of the truncated graph-total variation \cite{zhang2009some} on the inter-edges. It penalizes similar embeddings at the border between objects. Conscious that our embeddings are restricted to the unit sphere, we threshold this function for differences larger than $1$ (corresponding to a $60$ degree angle). In other words, $\psi(x)$ encourages vertices linked by an inter-edge to take embeddings with an euclidean distance of $1$, but does not push for a larger difference.

Note that any embeddings that are constant within objects, and with a difference of at least $1$ between adjacent objects, will have $0$ loss. The four-color theorem \cite{gonthier2008formal} tells us that it is always possible as long as the dimension of  our embedding is at least $3$. However, because embeddings are computed by the $\LCE$, borders which do not present recognizable geometric or radiometric configurations cannot be recovered by our method.% In fact, we know that the number of points which can be taken within the unit sphere with at least $60$ degree between them increases exponentially with the dimension.
%....................................................................
\subsubsection{Cross-Partition Weighting}
%...................................................................
The choice of $\mu_{i,j}$ plays a crucial role in the efficiency of the graph-structured contrastive loss. Although \ref{prop:p2} does imply \ref{prop:p1}, small errors in the former can have drastic consequences in the latter. Indeed, a single missed edge can erroneously fuse two large superpoints covering different objects. Therefore, we need to incorporate the induced partition's purity into the loss.

\cite{liu2018learning} introduced the segmentation-aware affinity loss (SEAL) implementing this idea. They propose weighting intra-edges as $1$, and inter-edges as $\mu_{i,j}=1+\vert S \mid - \vert S \setminus O_S \vert$ for $i$ and $j$ within the same superpoint $S$, with $O_S$ the \emph{majority-object}, \ie the object for which most points of $S$ belongs to. Although \cite{liu2018learning} boasts impressive results for superpixel oversegmentation, we were not able to extend this success within our framework. We believe this stems from three reasons: (i) all border edges of a superpoint are weighted identically regardless of their influence on the purity and the size of the interface; (ii) as soon as a superpoint no longer overlaps an object's border, its weight decreases dramatically to $1$, making the loss very unstable; (iii) \cite{liu2018learning} uses a different graph-based clustering\cite{liu2011entropy}.

\begin{figure}
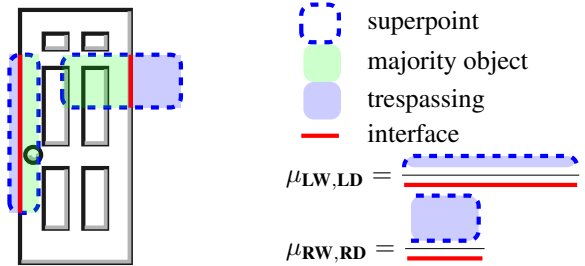

\include{door}
  \caption{Illustration of the cross-partition weighting strategy on a scene comprised of a door (\textbf{D}) and a wall (\textbf{W}). Two superpoints \textbf{L} (left) and \textbf{R} (right) overlap the door. 
  The superedge $(\textbf{LW},\textbf{LD})$(resp. $(\textbf{RW},\textbf{RD})$) represent the adjacency between the part of the left (resp. right) superpoint covering the wall and the part covering the door. With fewer trespassing points and a longer interface than $(\textbf{RW},\textbf{RD})$, the weights of the edges constituting $(\textbf{LW},\textbf{LD})$ are smaller.}
  \label{fig:gseal}
\end{figure}

To overcome these limitations, we introduce the cross-partition weighting strategy. We first compute the \emph{cross-segmentation graph} $\cG=(\cC,\cE)$, defined as the adjacency graph of the cross-partition $\cC$ of $C$ between the superpoints partition $\cS$ and the object partition $\cO$. In other words, $\cC$ is the set of connected components of the graph $G$ when all edges either between objects \emph{or} between superpoints are removed, and the \emph{super-edge} (\ie set of edges) $(U,V) \in \cE$ is the set of inter-edges of $\Einter$ between $U$ and $V$ in $\cC$:
\begin{align}\nonumber
\cC &= \Cur{O \cap S\mid O \in \cO, S \in \cS}\\\nonumber
\cE &= \Cur{\Cur{(i,j) \in \Pa{U \times V} \cap \Einter} \mid U,V \in \cC}.
\end{align}
We associate the following weight $\mu_{U,V}$ to each superedge $(U,V)$ and $\mu_{i,j}$ to each edge:
\begin{align}\nonumber
\mu_{U,V} =\;& 
\frac{\mu \min\Pa{\mid U \mid, \mid V \mid}}{\mid (U,V) \mid}  &\text{for}\; (U,V) \in \cE \\\nonumber
\mu_{i,j} =\;& \mu_{U,V}
  &\text{for all}\; (i,j) \in (U,V)
\end{align}
with $\mu$ a parameter of the model. Such weights simultaneously take into account the influence of the edges in the purity and the shape of the interfaces. Indeed, should an edge of the superedge $(U,V)$ be missed as a border, the superpoints $U$ and $V$ would be merged. 
Since $U$ and $V$ cover different objects (by definition of $\cE$), such a merger would induce at least $\min\Pa{\mid U \mid, \mid V \mid}$ vertices \emph{trespassing}, \ie not being in the majority-object of the merged superpoint.% over the majority object of $S$, 
%following the terminology of \cite{BENSHABAT2018}. %Another advantage of this approach is that it propagates the influence of the results of the non-differentiable graph partition step, penalizing configurations that causes the $\ell_0$ cut pursuit to miss border edges.
The weights are also divided by the number of edges constituting the interface between $U$ and $V$ in order to distribute evenly the penalty over the number of edges constituting an interface. This prevents long borders from being over-represented in the loss. See \figref{fig:gseal} for an illustration.%For example, the long border between $A\cap4-B\cap4$ in \figref{fig:gseal} only causes a small area to trespass on the boundary, and will be weighted less than short-boundary, high-impact $A\cap1-B\cap1$.
\setlength{\belowcaptionskip}{-4pt}
%....................................................................
\subsection {Cluster-Based Oversegmentation}
%....................................................................
\label{sec:cluster}
We also implemented a generalization of the method of \cite{JampaniSLYK18} to the 3D setting. The main advantage of this approach is that the loss can directly implement \ref{prop:p1} through the cross-entropy of the averaged semantic classes within superpoints. However, this approach remains hindered by its sensitivity to the superpoint initialization, and its inability to adapt the superpoints' size to the local complexity of the scene. Furthermore, as it bypasses \ref{prop:p3}, it produces superpoints with complicated contour.
%----------------------------------------------------------------------
\setlength\tabcolsep{3.2pt}
\begin{figure*}[!ht]
\begin{tabular}{c}
\begin{subfigure}[b]{1\textwidth}
\begin{tabular}{cccccc}
\includegraphics[width=.15\textwidth]{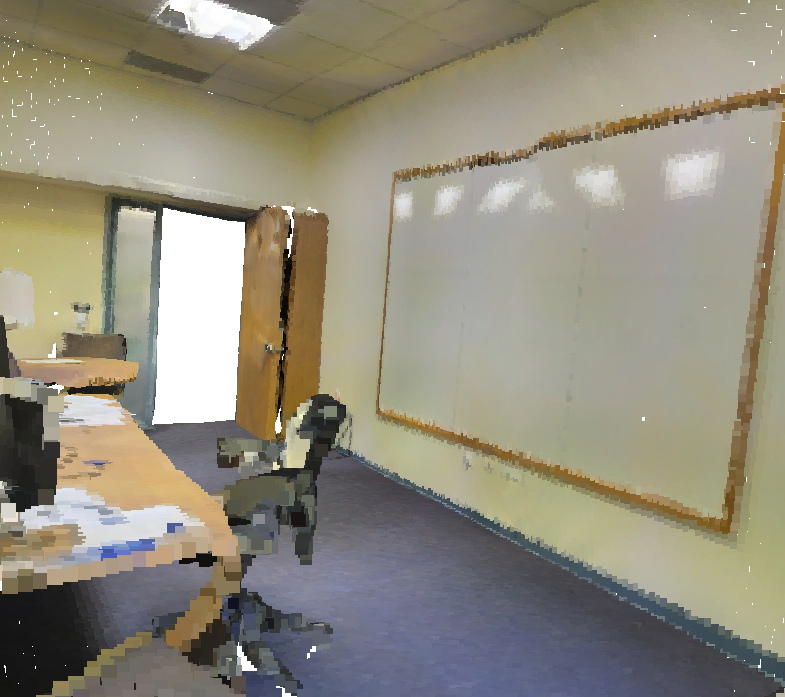}
&
\includegraphics[width=.15\textwidth]{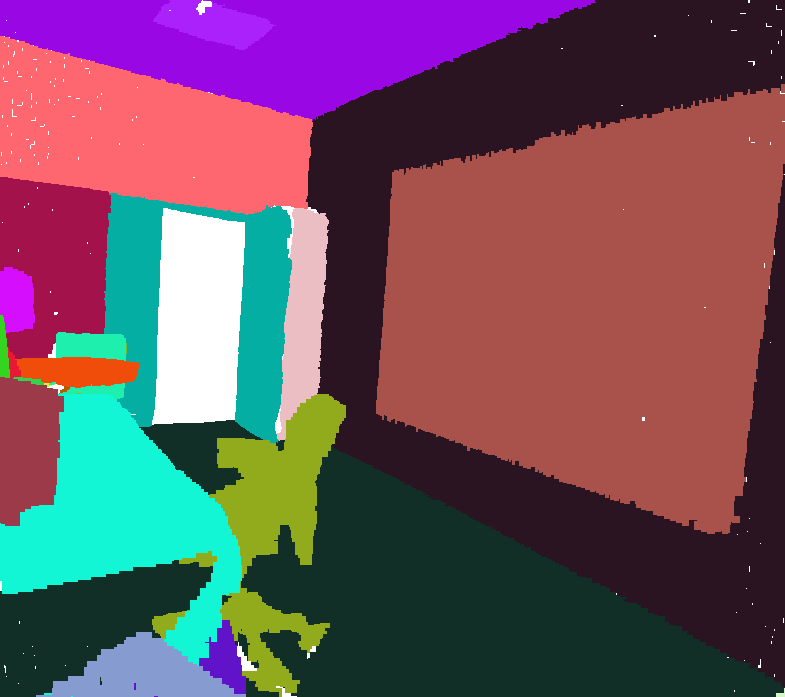}
&
\includegraphics[width=.15\textwidth]{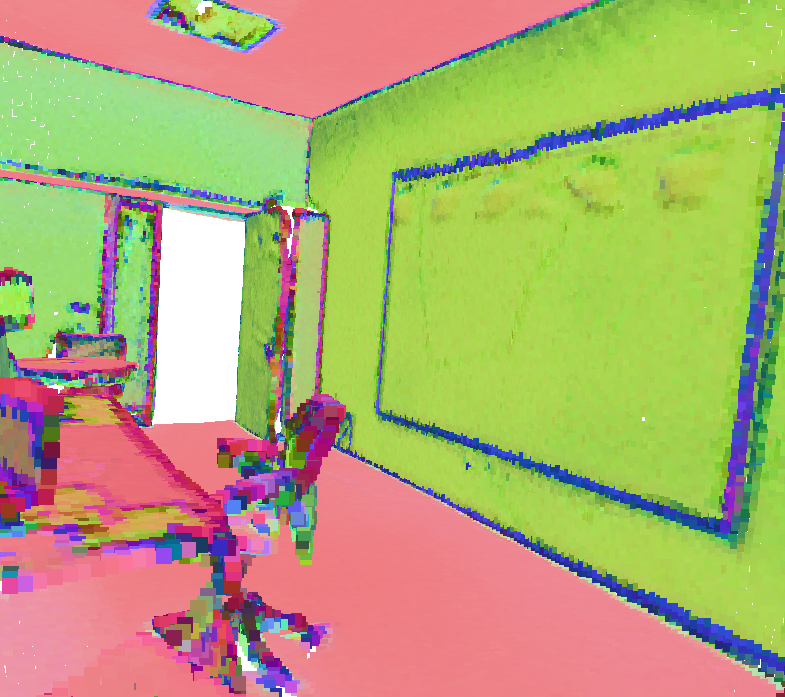}
&
\includegraphics[width=.15\textwidth]{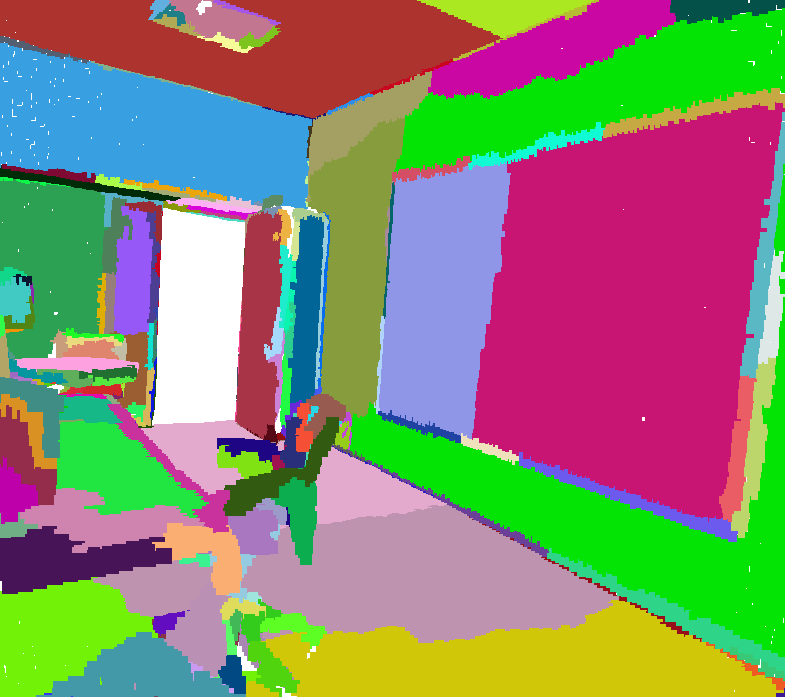}
&
\includegraphics[width=.15\textwidth]{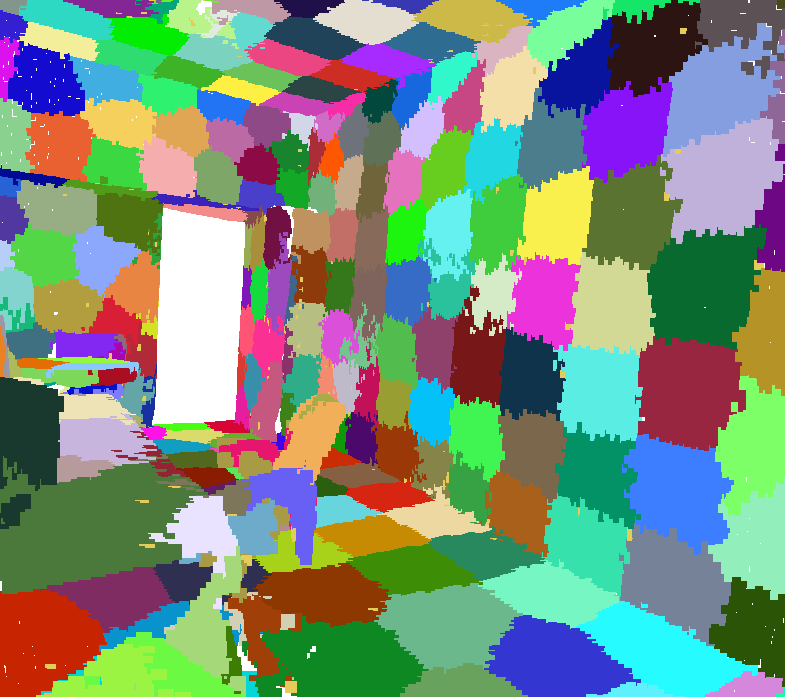}
&
\includegraphics[width=.15\textwidth]{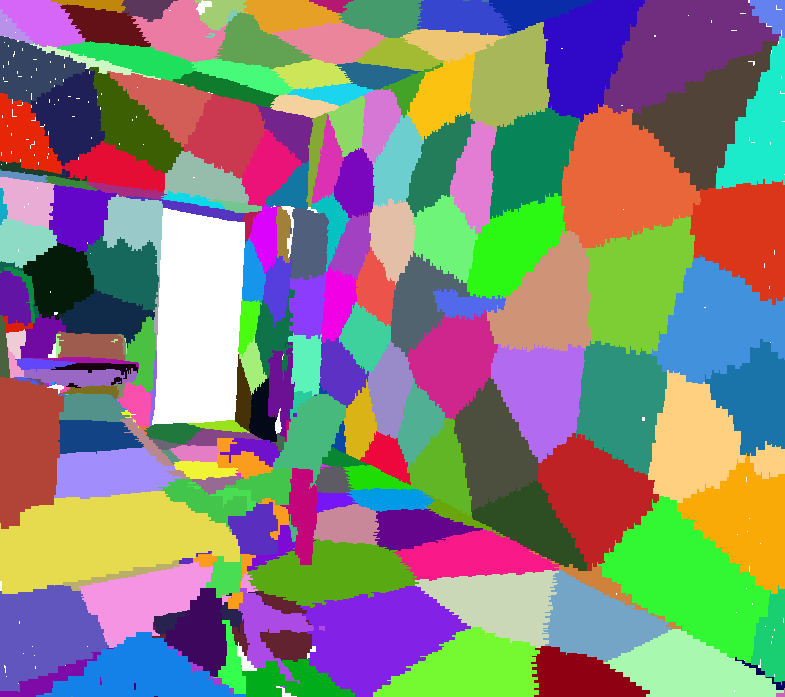}
\end{tabular}
\caption{S3DIS scene with 58 objects. Superpoint count : SSP 442, VCCS 436, Lin 423.}
\end{subfigure}\\
\begin{subfigure}[b]{1\textwidth}
\begin{tabular}{cccccc}
\includegraphics[width=.15\textwidth]{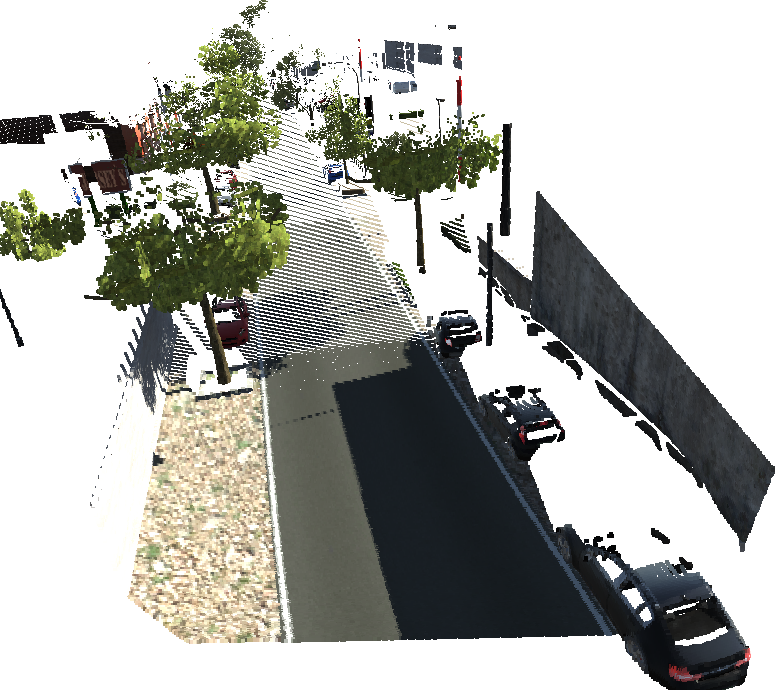}
&
\includegraphics[width=.15\textwidth]{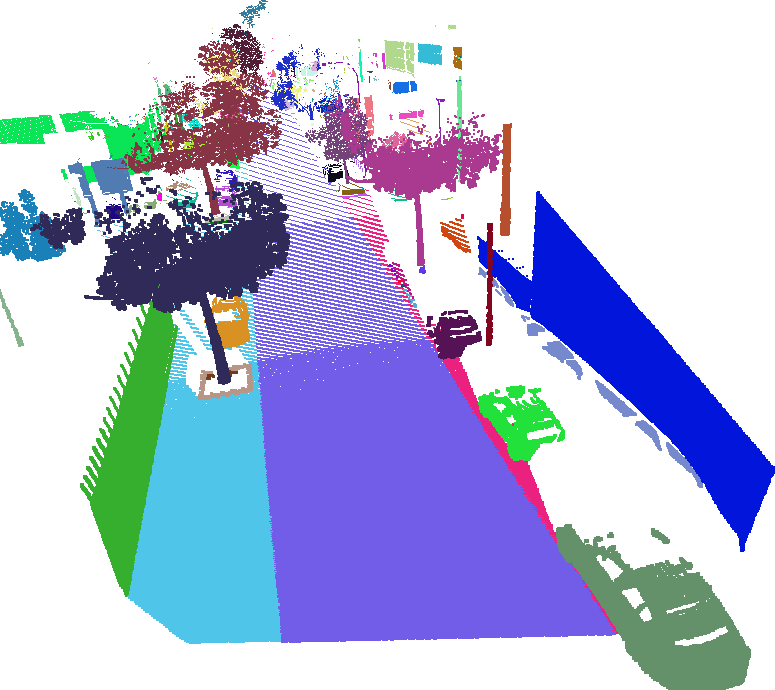}
&
\includegraphics[width=.15\textwidth]{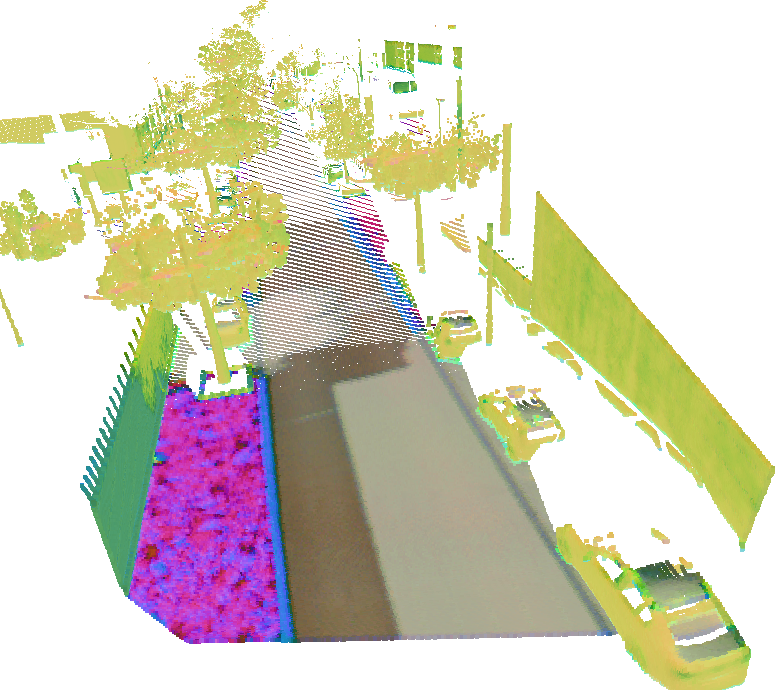}
&
\includegraphics[width=.15\textwidth]{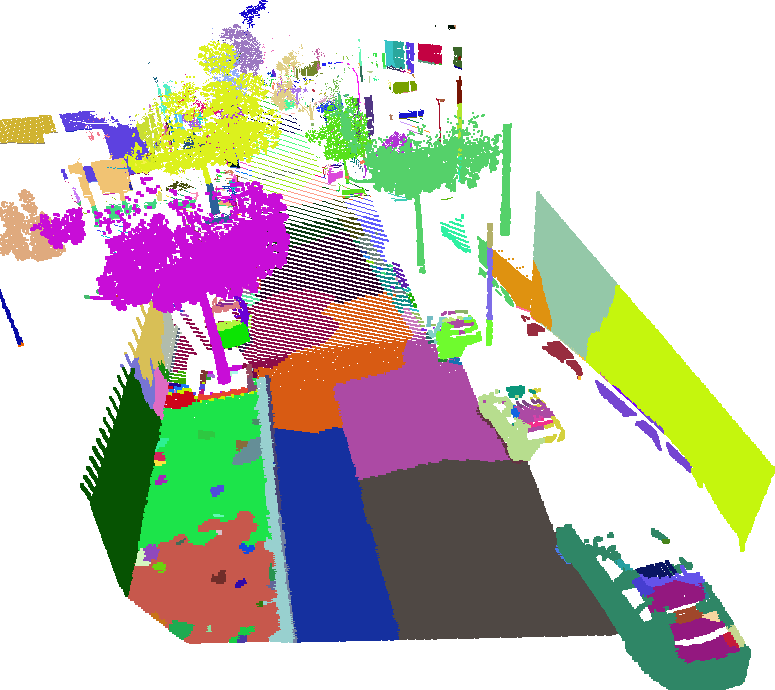}
&
\includegraphics[width=.15\textwidth]{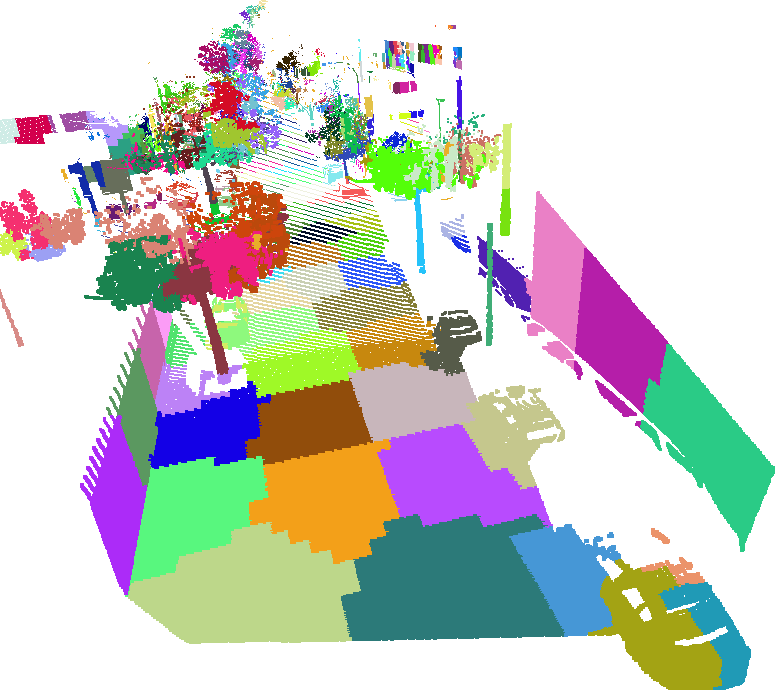}
&
\includegraphics[width=.15\textwidth]{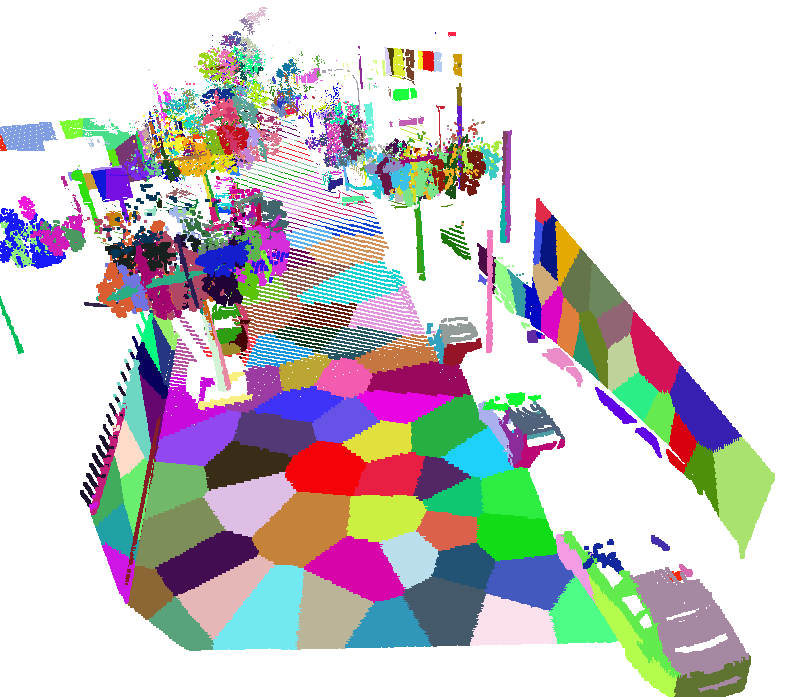}
\\
Input cloud&
Ground truth objects&
$\LCE$ embeddings&
SSP (ours)&
VCCS \cite{PaponASW13}&
Lin in \cite{LIN201839}
\end{tabular}
\caption{vKITTI scene with  $233$ objects. Superpoint count: SSP 420, VCCS 422, Lin 425.}
\end{subfigure}%\\
%\begin{subfigure}[b]{1\textwidth}
%\begin{tabular}{cccccc}
%\includegraphics[width=.15\textwidth]{./images/ini2.png}
%&
%\includegraphics[width=.15\textwidth]{./images/res2.png}
%&
%\includegraphics[width=.15\textwidth]{./images/tot2.png}
%&
%\includegraphics[width=.15\textwidth]{./images/obj2.png}
%&
%\includegraphics[width=.15\textwidth]{./images/obj2.png}
%&
%\includegraphics[width=.15\textwidth]{./images/obj2.png}
%\end{tabular}
%\caption{vKITTI dataset without color information}
%\end{subfigure}
\end{tabular}
\setlength{\belowcaptionskip}{-8pt}
\caption{Illustration of the oversegmentations of our framework, and from competing algorithms.}
\label{fig:illu_res}
\end{figure*}
\setlength{\belowcaptionskip}{0pt}
%.........................................................................
%....................................................................
\subsection {Implementation Details}
%....................................................................
\label{sec:details}
We use a modified version of the $\ell_0$-cut pursuit algorithm\footnote{\url{https://github.com/loicland/cut-pursuit}}\cite{landrieu2017cut}, with two main differences:
\begin{itemize}
\item to prevent the creation of many small superpoints in regions of high contrast, we merge components greedily with respect to the objective energy defined in \eqref{eq:mgp}, as long as they are smaller than a given threshold ;
\item we heuristically improved the forward step (8) from \cite{landrieu2017cut}, such that the regularization strength increases geometrically by a factor (of $0.7$) along the iterations. This helps improve the quality of the lower optima retrieved, and consequently the oversegmentation's.
\end{itemize}

To limit the size of the superpoints we concatenate to the points' embeddings their 3D coordinates in \eqref{eq:mgp} multiplied by a parameter $\aspat$, in the manner of \cite{achanta2012slic}. This determines the maximum size that superpoints can reach.

In all our experiments, we set $m$ the dimension of our embeddings to $4$. We choose a light architecture for the $\LCE$, with less than $15,000$ parameters. The exact network configurations for each dataset are detailed in the appendix. %==========================================================================
\section {Numerical Experiments}
%==========================================================================
%.........................................................................
\subsection {Datasets}
%.........................................................................
We evaluate our approach on two datasets of different natures. 
The first one is S3DIS \cite{Armeni16_s3dis}, composed of dense indoor scans of rooms in an office setting. The second one is vKITTI \cite{EngelmannKHL17_vkitty}, an outdoor dataset of urban scenes that mimics sparse LiDAR acquisitions.
Note that only S3DIS has individual object annotation. We consider the objects of vKITTI to be the connected components of the semantic labels in the adjacency graph $G$. For vKITTI, we consider the performance of our algorithm with and without color information.
Both datasets are large scale (close to $600$ million points for S3DIS and close to $15$ million for vKITTI). We subsample them using a regular grid of voxels (3cm wide for S3DIS and 5cm wide for vKITTI). In each voxel, we average the position and color of the contained points. This allows us to decrease the computation time and memory load.
%.........................................................................
\subsection {Point Cloud Oversegmentation}
%.........................................................................
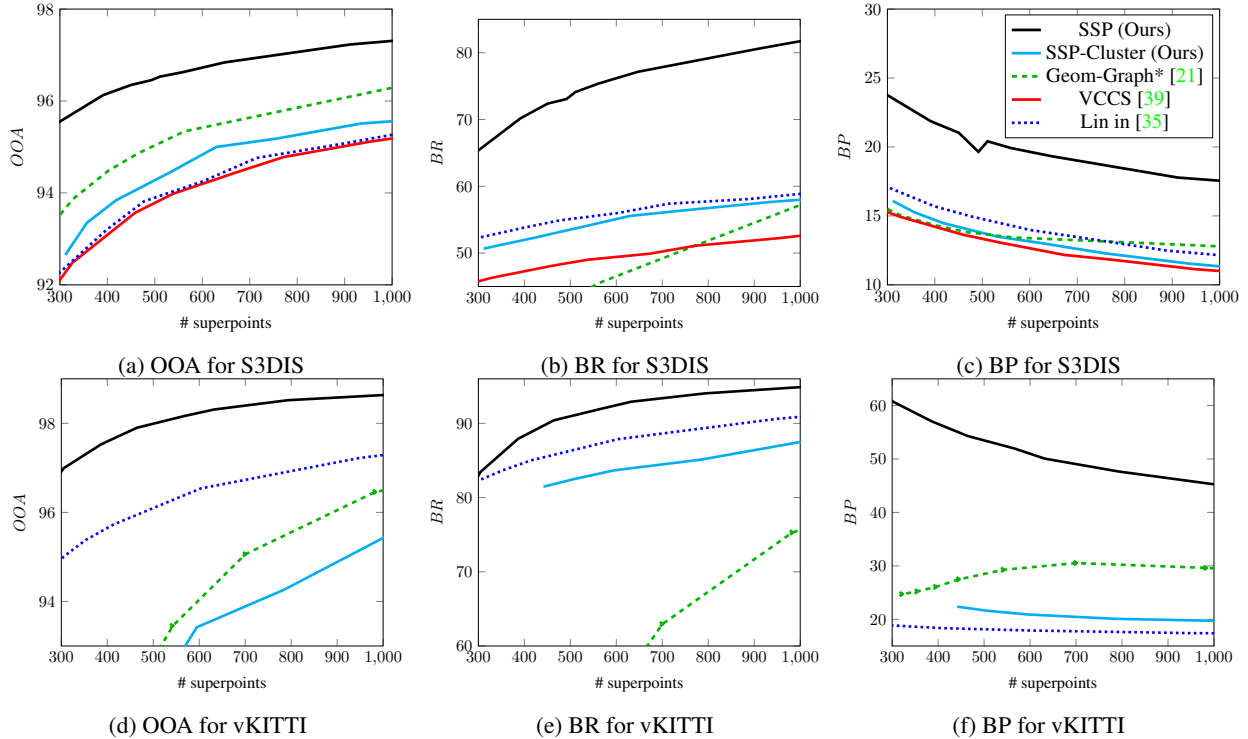
\begin{figure*}
 \input{spseg}
 \caption{Performance of the different algorithms on the 6-fold S3DIS dataset (\subref{fig:OA:S3DIS}, \subref{fig:BR:S3DIS}, \subref{fig:BP:S3DIS}), %the vKITTI dataset without color (\subref{fig:OA:VKITTI}) 
 and the 6-fold vKITTI dataset (\subref{fig:OA:VKITTIRGB},  \subref{fig:BR:VKITTIRGB}, \subref{fig:BP:VKITTIRGB}). The results of the method annotated with an asterix * have not been reported before. SSP-Cluster and VCCS are not represented for vKITTI for the sake of legibility as their performance is too low.}
 \label{fig:spseg}
\end{figure*}
\noindent
\textbf{Evaluation Metrics:} There are many standard metrics which assess the quality of point cloud oversegmentations with respect to properties \ref{prop:p1}, \ref{prop:p2}, and \ref{prop:p3}.
In particular, the Boundary Recall (BR) and Precision (BP) are used to evaluate the ability of the superpoints to adhere to, and not cross, object boundaries (\ref{prop:p2}, \ref{prop:p3}).
In the literature, these measures are defined with respect to \emph{boundary pixels} \cite{PaponASW13} or points \cite{LIN201839}. However, we argue that transition occurs \emph{between} points and not \emph{at} points for point clouds. Consequently, we define $\Einterp$ the set of predicted transition, \ie the subset of edges of $E$ that connect two points of $C$ in two different superpoints. These metrics are often given with respect to a tolerance, \ie the distance at which a predicted transition must take place from an actual object's border for the latter to be considered retrieved. We set this distance to $1$ edge, which leads us to define $\Eintere$ the set of inter-edges expanded to all directly adjacent edges in $E$:
$$\Eintere = \Cur{(i,j) \in E \mid \exists (i,k) \;\text{or}\; (j,k) \in \Einter}.$$
This allows us to define the boundary recall and precision with $1$ edge tolerance for a set of predicted transition $\Einterp$:
$$
BR = \frac{\mid\Einterp \cap \Eintere\mid}{\mid\Einter\mid},\;
BP = \frac{\mid\Einterp \cap \Eintere\mid}{\mid\Einterp\mid}.
$$
Since the end-goal of our point cloud oversegmentation framework is to provide useful superpoints for semantic segmentation, we define the \emph{Oracle Overall Accuracy} (OOA). To assess object purity \ref{prop:p1}, this metric characterizes the accuracy of the labeling that associates each superpoint $S$ of a segmentation $\cS$ with its majority ground-truth label. Formally, let $l \in \cK^C$ be the semantic labels of each point within a set of classes $\cK$, we define the OOA of a point cloud segmentation $\cS$ as:
\begin{align}\nonumber
    &\loracle(S)=\text{mode}\Cur{l_i \mid i \in S}\\\nonumber
    &OOA= \frac1{\mid C \mid} \sum_{S \in \cS}\sum_{i \in S} \Bra{l_i=\loracle(S)},
\end{align}
with $[x=y]$ the function equal to $1$ if $x=y$ and $0$ otherwise. Note that the OOA is closely related to the ASA \cite{liu2011entropy}, but consider the majority labels of all points within a superpixel rather than the label of the objects with most overlap. In this sense, it is a tighter upper bound to the achievable accuracy of a superpoint-based semantic classification algorithm using $\cS$. This metric is also more fair than the undersegmentation error \cite{levinshtein2009turbopixels} for other methods such as \cite{guinard2017weakly}, or our cluster-based approach, as they do not try to retrieve objects directly, but rather regions of $C$ with homogeneous semantic labeling.

\noindent
\textbf{Competing algorithms:} %We evaluate the performance of our model against different approaches.
We denote by \textbf{SSP} (Supervized SuperPoint) our method when using LPE to learn point embeddings and then derive the superpoints using the graph-based methods described in \secref{sec:graph}, and \textbf{SSP-Cluster} when using the cluster-based method defined in \secref{sec:cluster} instead.
We first assess the benefit of learning embeddings by comparing our results to those of \cite{guinard2017weakly}, dubbed here \textbf{Geom-Graph}. This method computes superpoints by solving the generalized minimal partition problem as well, but with handcrafted geometric features in place of our learned embeddings. We illustrate in \figref{fig:spseg} the oversegmentations produced by our approach and two state-of-the-art algorithms: \textbf{VCCS} \cite{PaponASW13} and the work of Lin in \cite{LIN201839}.

We observe that our approach significantly outperforms the other approaches on all metrics. In particular, we remark that \textbf{SSP} only requires under $350$ superpoints to reach a performance comparable with \textbf{VCCS} with over $1,800$ superpoints on S3DIS. Furthermore, the quality of the border is unmatched in our range of superpoints. The improvement is less significant on vKITTI, which could be due to the difficulty of constructing an adjacency graph on such a sparse acquisition. The performance is degraded further without color information, as some transition are not predictable with purely from the geometry.
\textbf{Geom-Graph} performs well on the accuracy, but not on the boundary. This is expected as the handcrafted geometric features cannot detect some borders, such as adjacent walls. \textbf{SSP-Cluster} performs better than the unsupervized cluster-based method of Lin \etal, but still suffer from the typical limitations of clustering methods, such as sensitivity to initialization.% or intricate borders.

In terms of computational speed, the embeddings can be computed very efficiently in parallel on a GPU with over $3$ million embeddings per second on a 1080Ti GPU. The bottleneck remains solving the graph partition problem in \eqref{eq:mgp}, which can process around $100,000$ points per second.
%.........................................................................
\subsection {Semantic Segmentation}
%.........................................................................
In \tabref{tab:s3dis:semseg} and \tabref{tab:vkitti:segsem}, we show how our point cloud oversegmentation framework can be successfully used by the superpoint-based semantic segmentation technique of \cite{landrieu2017large}\footnote{\label{foot:spg}\url{https://github.com/loicland/superpoint-graph}} (\textbf{SPG}). We replace the unsupervized superpoint computation with our best-performing approach, \textbf{SSP}. We evaluate the resulting semantic segmentation using standard classification metrics: overall accuracy (OA), mean per-class accuracy (mAcc) and mean per-class intersection-over-union (mIOU). We observe a significant increase in the performance of \textbf{SPG}, beating  concurrent methods on both datasets. In particular, we observe that our method allows for better retrieval of small objects (see detailed IoU in the appendix), which translates into much better per-class metrics, although the overall accuracy is not necessarily better than the latest state-of-the-art algorithms. 
%.........................................................................
\subsection{Ablation Study}
%.........................................................................
In \tabref{tab:ablation}, we present an ablation study to empirically justify some of our design choices. To make things more legible, we present the increase/decrease of the 3 performance metrics at $500$ superpoints (linearly interpolated) of alternative methods compared to ours, on the first cross-validation fold of the S3DIS dataset. In particular we
present \textbf{Prop-weight}, an alternative version in which the cross-partition weighting is replaced by a simple inversely-proportional weighting of the inter/intra edges. Predictably, this method gives lesser results as the edges are not weighted according to their influence in the partition. However, since the weights of the intra-edge are proportionally higher, the border precision is improved. We implemented the weights of the segmentation-aware affinity loss of \cite{liu2018learning} as well for method \textbf{SEAL-weights}, with comparable results to the \textbf{Prop-weight}. In \textbf{+TV-TV}, we replace our choice of function $\phi$ and $\psi$ in the loss by respectively $\mid \cdot \mid$ and $-\mid \cdot \mid$, so that our loss is closer to the pairwise affinity loss used by \cite{engelmann2018} (but still structured by the graph). However, this approach wouldn't give meaningful partition as the intra-edge term conflicts with the constraint that the embeddings are constrained on the sphere. Removing this restriction leads the collapse of the embeddings around $0$. 
We also tried to stack the $\LCE$ in layers, using or not a residual structure comparable to the one used in \cite{he2016deep} to increase their receptive fields (more details are given in the appendix). The best results were achieved with two layers: \textbf{2-Layers} and \textbf{2-Residuals}. However, we observe that when compared with $\LCE$ of a similar number of parameters, the gains are insignificant if not null. We conclude that to embed points in order to detect borders, a small receptive field with a shallow architecture is sufficient.
%
\input{tables}
\begin{table}
\begin{center}
\begin{tabular}{c|cccc}
\bf{Method} & \# parameters & OOA  & BR & BP \\\hline
\bf{Best} & 13,816& 96.2 & 73.3 & 22.1 \\
\bf{Prop-weights} &13,816 & -2.6 & -12.2& +10.4\\
\bf{SEAL-weights} & 13,816& -1.3& -11.3& +3.8 \\
%{+TV-TV} & 13,816 & & & \\
\bf{2-Layers} & 14,688 &-0.1 & -0.7 & -0.3 \\
\bf{2-Residuals}& 14,688 & +0.0 & -0.2 & -0.7 
\end{tabular}
\end{center}
\caption{Impact of some of our design choice on S3DIS. \textbf{Best} is the \textbf{SSP} method with cross-partition weights.}
\label{tab:ablation}
\end{table}
%=========================================================================
\section{Conclusion}
%=========================================================================
%\vspace{-2mm}
In this paper, we presented the first supervized 3D point cloud oversegmentation framework. Using a simple point embedding network and a new graph-structured loss function, we were able to achieve significant improvements compared to the state-of-the-art of point cloud oversegmentation. When combined with a superpoint-based semantic segmentation method, our method sets a new state-of-the-art of semantic segmentation as well. A video illustration is accessible at \url{https://youtu.be/bKxU03tjLJ4}.
The source code will be made available to the community as well as trained networks in an update to the superpoint-graph repository\footnoteref{foot:spg}. Future work will focus on improving the solving method for the generalized minimum minimal partition problem to better handle spherically-bounded variables, and to improve its computational performance. %From a more general perspective, we believe that using deep networks to learn parameters and input of optimization problem is a promising venue.
\clearpage
\balance
\small
\bibliographystyle{ieee}
%\bibliography{mybib}

%\clearpage
%\appendix
%\section*{Appendix} 
%\input{supplementary}

\ARXIV{
\clearpage
\appendix
\section*{Supplementary Material} 
\input{supplementary}
}{}

\end{document}

%% file: figure_STN.tex
%auto-ignore
\begin{tikzpicture}
\tikzstyle{input}=[draw = none, fill = none, scale = 0.1]
\tikzstyle{operator}=[draw = black, very thick]
\tikzstyle{pointfeature}=[draw = black,  ->, very thick]
\tikzstyle{setfeature}=[draw = black, dotted, ->, very thick]
\tikzstyle{operator}=[circle, draw = black, very thick, inner sep=0pt,
  text width=5mm,align=center]
  
\tikzstyle{crossed}=[circle, draw = black, very thick, inner sep=0pt,
  text width=5mm,align=center,
    path picture={
        \draw
            (path picture bounding box.south east) -- (path picture bounding box.north west)
            (path picture bounding box.south west) -- (path picture bounding box.north east);
        }
    ]
%\node at (0,0.5) [input,minimum width=.25cm,minimum height=1.5cm,label={left:$\Pi$}] (setfeat){};
%\node at (1.2,2) [input,minimum width=.25cm,minimum height=.25cm,label={left:$\pi$}] (pointfeat){};
\coordinate (setfeat) at (0,2.0);
\coordinate (pointfeat) at (0.0,0.0);

\coordinate (setfeatout) at (7.5,2.0);
\coordinate (pointfeatout) at (7.5,0.0);

\node at (setfeat) [input, label={left:$P_i$}] (setfeatname){};
\node at (pointfeat) [input,label={[yshift = +1mm]left:$p_i$}] (pointfeatname){};

\node at (setfeatout) [input, label={left:$\Pit$}] (setfeatname){};
\node at (pointfeatout) [input,label={[yshift = +1mm]left:$\pit$}] (pointfeatname){};

\node at (1.2,2.0) [operator,circle split] (minus){};
\node at (2.5,0.0) [operator] (z){z};
\node at (2.5,1.0) [operator] (diam){r};
\node at (3.5,2.0) [operator, forbidden sign] (div){};
\node at (5,1.0) [operator, rectangle, text width=7mm, inner sep=2pt] (ptn){PTN};
\node at (6,2.0) [crossed] (prod){};

\draw [setfeature] (setfeat) -- (minus) node [pos=0.4, above, fill=none] {\tiny $k \times 3$};
\draw [pointfeature] (pointfeat) -- (1,0.0) node [pos=0.2, above, fill=none] {\tiny $3$} -- (z);
\draw [pointfeature]  (z) -- (3,0.0) node [pos=0.7, above, fill=none] {\tiny $1$} -- (6,0.0) node [pos=0.2, above, fill=none] {\tiny $2$} -- (7,0.0) node [pos=0.4, above, fill=none] {\tiny $6$};
\draw [pointfeature] (1.2,0.0) -- (minus);
\draw [setfeature] (minus) -- (div);
\draw [setfeature] (minus) -- (1.7,2.0) -- (1.7,1.0) -- (diam);
\draw [pointfeature] (3.2,1.0) -- (3.2,0.0);
\draw [pointfeature] (diam) -- (3.5,1.0) node [pos=0.2, above, fill=none] {\tiny $1$} -- (div);
\draw [setfeature] (div) -- (prod);
\draw [setfeature] (4,2.0) -- (4,1.0) -- (ptn);
\draw [pointfeature] (ptn) -- (6,1.0) node [pos=0.3, above, fill=none] {\tiny $4$} -- (prod);
\draw [pointfeature] (6,1.0) -- (6,0.0);
\draw [setfeature] (prod) -- (7,2.0) node [pos=0.4, above, fill=none] {\tiny $k \times 3$};
\end{tikzpicture}

%% file: figure_RCE.tex
%auto-ignore
    \begin{tikzpicture}
 
 \tikzstyle{plus}=[circle, draw = black, very thick, inner sep=0pt,
  text width=5mm,align=center,
    path picture={
        \draw
            (path picture bounding box.south) -- (path picture bounding box.north)
            (path picture bounding box.west) -- (path picture bounding box.east);
        }
    ]
 
\tikzstyle{input}=[draw = none, scale = 0.1]
\tikzstyle{operator}=[draw = black, very thick]
\tikzstyle{pointfeature}=[draw = black,  ->, very thick]
\tikzstyle{setfeature}=[draw = black, dotted, ->, very thick]
 \tikzstyle{operator}=[circle, draw = black, very thick, inner sep=0pt,
  text width=3mm,align=center]
  
\coordinate (setfeat) at (0,0);
\coordinate (pointfeat) at (3.5,-1.0) ;
%\coordinate (embed) at (7.5,0.0);

 \node at (setfeat) [input,minimum width=.25cm,minimum height=1.5cm,label={left:$X_i$}] (setfeatname){};
 \node at (pointfeat) [input,minimum width=.25cm,minimum height=1.5cm,label={above left:$x_i$}] (pointfeatname){};

\node at (1.15,0) [operator, rectangle, text width=9mm, inner sep=2pt] (mlp1){$\MLP_1$};

\node at (2.5,0) [draw = white, text width=12mm, inner sep=00pt, rotate = -90] (max){\small maxpool};
\draw [very thick] (2.3,-0.9) -- (3.2,0) -- (2.3,0.9) -- cycle;

\node at (4.5,0) [operator, rectangle, text width=9mm, inner sep=2pt] (mlp2){$\MLP_2$};

%\node[plus] at (5.5,0)(plus) {};

\node at (6.1,0) [operator, rectangle, text width=5mm, inner sep=2pt] (l2){L2};

%\node at (7.2,0) [input,minimum width=.25cm,minimum height=1.5cm] (emb){$e_i$};

\coordinate (emb) at (7.2,0.0);
 \node at (emb) [input,minimum width=.25cm,minimum height=1.5cm,label={right:$e_i$}] (embname){};

\draw [setfeature] (setfeat) -- (mlp1);

\draw [setfeature] (mlp1) -- (2.3,0);

\draw [pointfeature] (3.2,0) -- (mlp2);

\draw [pointfeature] (mlp2) -- (l2);

\draw [pointfeature] (l2) -- (emb);

\draw [pointfeature] (pointfeat) -- (3.5,0);

%\draw [pointfeature] (embed) -- (plus);

    \end{tikzpicture}

%% file: door.tex
%auto-ignore
\begin{tabular}{ll}
%\begin{subfigure}[t]{0.25\textwidth}
%\parbox[t]{1\textwidth}{
\begin{minipage}[b][.15\textheight][s]{.20\textwidth}
\begin{tikzpicture}
    \node[anchor=south west,inner sep=0] (image) at (0,0) {\includegraphics[width=1\textwidth]{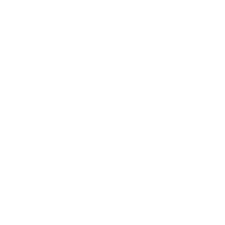}};
    \begin{scope}[x={(image.south east)},y={(image.north west)}]
     \tikzstyle{door}=[draw=black!90!white, ultra thick]
     \tikzstyle{shade}=[draw=black!30!white, line width = 0.5mm]
     %\draw[draw=none, fill = white] (00,0) rectangle (1,1);
        \draw[door] (0.09,0.01) rectangle (0.51,0.97);
        \draw[shade] (0.095,0.015) -- (0.505,0.015) -- (0.505,0.965);
        
        \draw[door] (0.18,0.14) rectangle (0.27,0.37);
        \draw[shade] (0.185,0.145) -- (0.265,0.145) -- (0.265,0.365);
        
        \draw[door] (0.33,0.14) rectangle (0.42,0.37);
        \draw[shade] (0.335,0.145) -- (0.415,0.145) -- (0.415,0.365);
        
        \draw[door] (0.18,0.46) rectangle (0.27,0.75);
        \draw[shade] (0.185,0.465) -- (0.265,0.465) -- (0.265,0.745);
        
        \draw[door] (0.33,0.46) rectangle (0.42,0.75);
        \draw[shade] (0.335,0.465) -- (0.415,0.465) -- (0.415,0.745);
        
        \draw[door] (0.18,0.82) rectangle (0.27,0.88);
        \draw[shade] (0.185,0.825) -- (0.265,0.825) -- (0.265,0.875);
        
        \draw[door] (0.33,0.82) rectangle (0.42,0.88);
        \draw[shade] (0.335,0.825) -- (0.415,0.825) -- (0.415,0.875);
        
        %\draw[shade] (0.345,0.415) circle (3pt);
        %\draw[shade] (0.345,0.415) arc (-90:0:2) ;
        \draw[shade] (0.128,0.418) arc (-180:30:0.015) ;
        \draw[door] (0.14,0.42) circle (3pt);
    
        \draw[blue,ultra thick,rounded corners, dashed] (0.25,0.6) rectangle (0.7,0.8);
        \fill[blue,draw=none, opacity=0.2] (0.5,0.6) rectangle (0.7,0.8);
        \fill[green,draw=none, opacity=0.2] (0.25,0.6) rectangle (0.5,0.8);
        
        \draw[blue,ultra thick,rounded corners, dashed] (0.05,0.2) rectangle (0.16,0.8);
        \fill[blue,draw=none, opacity=0.2] (0.1,0.2) rectangle (0.05,0.8);
        \fill[green,draw=none, opacity=0.2] (0.1,0.2) rectangle (0.16,0.8);
        
        \draw[red,ultra thick] (0.09,0.2) -- (0.09,0.8);
        
        \draw[red,ultra thick] (0.51,0.6) -- (0.51,0.8);

    \end{scope}
\end{tikzpicture}
%}
\end{minipage}
&
\begin{minipage}[b][.15\textheight][s]{.25\textwidth}
%\begin{subfigure}[t]{0.2\textwidth}
%\parbox[t]{.25\textwidth}{
%\begin{minipage}[t]{0.25\linewidth}
\begin{tabular}{rl}
\begin{tikzpicture}[baseline={([yshift=-.5ex]current bounding box.center)}]
  \draw[draw=blue,fill=none,ultra thick,rounded corners, dashed] (0.0,0.0) rectangle (0.5,0.5);
 \end{tikzpicture}& superpoint\\
\begin{tikzpicture}[baseline={([yshift=-.5ex]current bounding box.center)}]
  \draw[draw=none,fill=green,fill opacity=0.2,ultra thick,rounded corners, dashed] (0.0,0.0) rectangle (0.5,0.5);
 \end{tikzpicture}& majority object\\
 \begin{tikzpicture}[baseline={([yshift=-.5ex]current bounding box.center)}]
  \fill[draw=none,fill=blue,fill opacity=0.2,ultra thick,rounded corners, dashed] (0.0,0.0) rectangle (0.5,0.5);
 \end{tikzpicture}&trespassing
 \\
 \begin{tikzpicture}[baseline={([yshift=-.5ex]current bounding box.center)}]
  \draw[red,ultra thick] (0.0,0.0) -- (0.5,0.0);
 \end{tikzpicture}& interface
 \end{tabular}\\
%}
$\mu_{\textbf{LW},\textbf{LD}} =$
 $
%\frac
{
\begin{tikzpicture}[baseline={([yshift=-.5ex]current bounding box.center)}]
 \fill[draw = none, fill = blue!20!white, rounded corners] (0,0) rectangle (2.3,0.15);
  \draw[draw = blue, ultra thick,rounded corners, dashed] (0,0) -- (0,0.15) --  (2.3,0.15) -- (2.3,0.0);
  %\draw[draw = blue, ultra thick, dashed] (0,0) -- (2.3,0.0);
      \draw[draw = none] (0,0) -- (0,-0.1);
 \end{tikzpicture}
 }
 \oover
 {
 \begin{tikzpicture}[baseline={([yshift=-.5ex]current bounding box.center)}]
  \draw[red,ultra thick] (0.0,0.0) -- (2.3,0.0);
 \end{tikzpicture}
 }
 $
 \\
$\mu_{\textbf{RW},\textbf{RD}} = $
$
%\frac
{
\begin{tikzpicture}[baseline={([yshift=-.5ex]current bounding box.center)}]
 \fill[draw = none, fill = blue!20!white, rounded corners] (0,0) rectangle (.9,0.6);
  \draw[draw = blue, ultra thick,rounded corners, dashed] (0,0) -- (.9,0) --  (.9,0.6) -- (0,0.6);
  %\draw[draw = blue, ultra thick, dashed] (0,0) -- (0,0.6);
  \draw[draw = none] (0,0) -- (0,-0.1);
 \end{tikzpicture}
 }
 \oover
 {
 \begin{tikzpicture}[baseline={([yshift=-.5ex]current bounding box.center)}]
  \draw[red,ultra thick] (0.0,0.0) -- (1,0.0);
 \end{tikzpicture}
 }
 $
\end{minipage}
\end{tabular}

%% file: spseg.tex
%auto-ignore
  \pgfplotsset{
% override style for non-boxed plots
    % which is the case for both sub-plots
    every non boxed x axis/.style={} 
}

\setlength{\belowcaptionskip}{-5pt}
\setlength{\abovecaptionskip}{-5pt}
\pgfplotstableread{./data/ptn_s3dis.txt}{\ptnstanford}
\pgfplotstableread{./data/cvpr_s3dis.txt}{\cvprstanford}
\pgfplotstableread{./data/isprs_s3dis.txt}{\isprsstanford}
\pgfplotstableread{./data/geof_s3dis.txt}{\geofstanford}
\pgfplotstableread{./data/slic_s3dis.txt}{\slicstanford}

\pgfplotstableread{./data/ptn_vkitticolor.txt}{\ptnvkitti}
\pgfplotstableread{./data/ptn_vkitticolor2.txt}{\ptnvkittid}
\pgfplotstableread{./data/cvpr_vkitti.txt}{\cvprkitti}
\pgfplotstableread{./data/isprs_vkitti.txt}{\isprskitti}
\pgfplotstableread{./data/geof_vkitti.txt}{\geofkitti}
\pgfplotstableread{./data/slic_vkitti.txt}{\slicvkitti}

\pgfplotstableread{./data/ptn_vkittinocolor.txt}{\ptnvkittinocolor}
\pgfplotstableread{./data/isprs_vkitti.txt}{\isprskittinocolor}
\pgfplotstableread{./data/geof_vkittinocolor.txt}{\geofkittinocolor}

 \tikzstyle{ptn}=[black, ultra thick]
 \tikzstyle{geof}=[green!70!black, ultra thick, dashed]
  \tikzstyle{slic}=[cyan, ultra thick]
 \tikzstyle{cvpr}=[red, ultra thick]
 \tikzstyle{isprs}=[blue, ultra thick, dotted]
 \setlength\tabcolsep{0pt}
\begin{center}
\begin{tabular}{ccc}
%\begin{multirow}{2}{*}{
 
  %}
  %&
  \setcounter{subfigure}{0}
  \begin{subfigure}[b]{.32\textwidth} %\centering
  \resizebox{1\textwidth}{!}{
    \begin{tikzpicture}
      \begin{axis}
        [xmin = 300, xmax = 1000, xlabel={\# superpoints}, ylabel={$OOA$} , ymin = 92, ymax = 98, y label style={at={(+0.05,0.5)}}]
        \addplot [ptn] table [x = {N}, y = {ASA}] {\ptnstanford};
        \addplot [slic] table [x = {N}, y = {ASA}] {\slicstanford};
        \addplot [geof] table [x = {N}, y = {ASA}] {\geofstanford};
        \addplot [cvpr] table [x = {N}, y = {ASA}] {\cvprstanford};
        \addplot [isprs] table [x = {N}, y = {ASA}] {\isprsstanford};
      \end{axis}
    \end{tikzpicture}
    }
    \caption{OOA for S3DIS}
    \label{fig:OA:S3DIS}
  \end{subfigure}&
   
\begin{subfigure}[b]{0.31\textwidth} %\centering
\resizebox{1\textwidth}{!}{
    \begin{tikzpicture}
      \begin{axis}
        [xmin = 300, xmax = 1000, xlabel={\# superpoints}, ylabel={$BR$}, ymin = 45, ymax = 85, y label style={at={(+0.05,0.5)}}]
        \addplot [ptn] table [x = {N}, y = {BR}] {\ptnstanford};
        \addplot [slic] table [x = {N}, y = {BR}] {\slicstanford};
       \addplot [geof] table [x = {N}, y = {BR}] {\geofstanford};
           \addplot [cvpr] table [x = {N}, y = {BR}] {\cvprstanford};
        \addplot [isprs] table [x = {N}, y = {BR}] {\isprsstanford};
      \end{axis}
    \end{tikzpicture}
    }
       \caption{BR for S3DIS}
       \label{fig:BR:S3DIS}
  \end{subfigure} 
  &
\begin{subfigure}[b]{.32\textwidth} %\centering
\resizebox{1\textwidth}{!}{
    \begin{tikzpicture}
      \begin{axis}
        [xmin = 300, xmax = 1000, xlabel={\# superpoints}, ylabel={$BP$} , ymin = 10, ymax = 30, y label style={at={(+0.05,0.5)}}]

        \addplot [ptn] table [x = {N}, y = {BP}] {\ptnstanford};
        \addplot [slic] table [x = {N}, y = {BP}] {\slicstanford};
        \addplot [geof] table [x = {N}, y = {BP}] {\geofstanford};
        \addplot [cvpr] table [x = {N}, y = {BP}] {\cvprstanford};
        \addplot [isprs] table [x = {N}, y = {BP}] {\isprsstanford};
        \addlegendentry{\large SSP (Ours)}
        \addlegendentry{\large SSP-Cluster (Ours)}
        \addlegendentry{\large Geom-Graph* \cite{guinard2017weakly}}
        \addlegendentry{\large VCCS \cite{PaponASW13}}
        \addlegendentry{\large Lin in \cite{LIN201839}}
      \end{axis}
    \end{tikzpicture}
     }
      \caption{BP for S3DIS}
      \label{fig:BP:S3DIS}
  \end{subfigure}
  \\
  
 %  \begin{subfigure}[b]{.25\textwidth%} %\centering
 % \resizebox{1\textwidth}{!}{
   % \begin{tikzpicture}
     % \begin{axis}
       % [xmin = 300, xmax = 1000, %xlabel={\# superpoints}, %ylabel={BR} , ymin = 50, ymax %= 95, y label %style={at={(+0.05,0.5)}}]
       % \addplot [ptn] table [x = %{N}, y = {BR}] %{\ptnvkittinocolor};
        %\addplot [cvpr] table [x = {N}, y = {ASA}] {\cvprstanford};
       % \addplot [isprs] table [x = %{N}, y = {BR}] %{\isprskittinocolor};
       % \addplot [geof, mark=x] table %[x = {N}, y = {BR}] %{\geofkittinocolor};
  %    \end{axis}
  %  \end{tikzpicture}
  %  }
  %  \caption{BR for vKITTI no color}
  %  \label{fig:OA:VKITTI}
  %\end{subfigure}
  %&
  \if 1 0
 \begin{subfigure}[b]{.32\textwidth} %\centering
 \resizebox{1\textwidth}{!}{
\begin{tikzpicture}

\begin{groupplot}[
    group style={
        group name=my fancy plots,
       group size=1 by 2,
       xticklabels at=edge bottom,
       vertical sep=0pt
    },
    width=5.5cm,
    xmin=300, xmax=1000
]
\nextgroupplot[ymin=95,ymax=100,
               ytick={96,98,99,100},
               axis x line=top, 
               axis y discontinuity=parallel,
               height=3.5cm]
 %\addplot [cvpr] table [x = {N}, y = {ASA}] {\cvprkitti};
\addplot [isprs] table [x = {N}, y = {ASA}] {\isprskitti};        
\addplot [black] {99+0.001*x};

\nextgroupplot[ymin=80,ymax=87,
               ytick={80,85},
               axis x line=bottom,
               height=3.0cm]
\addplot [cvpr] table [x = {N}, y = {ASA}] {\cvprkitti};
\addplot [isprs] {99 + x * 0.001};
%\addplot [isprs] table [x = {N}, y = {ASA}] {\isprskitti};  
   %width=3.5cm,
  % xmin = 300, xmax = 1000,
   %xlabel={\# superpoints}, ylabel={$OOA$},
  % y label style={at={(+0.05,0.5)}}
  %  ]

%\nextgroupplot[ymin=95,ymax=100,
 %             ytick={95,100},
  %            axis x line=top, 
  %             axis y discontinuity=parallel,
  %             height=5cm]
  %             
  %\addplot [cvpr] table [x = {N}, y = {ASA}] {\cvprkitti}; %\addplot [isprs] table [x = {N}, y = {ASA}] {\isprskitti};
  % 
   %
  % \nextgroupplot[ymin=80,ymax=90,
   %            ytick={80,85},
   %            axis x line=bottom,
    %           height=2cm]
   %%\addplot [cvpr] table [x = {N}, y = {ASA}] {\cvprkitti}; %\addplot [isprs] table [x = {N}, y = {ASA}] %{\isprskitti}; 
\end{groupplot}
\end{tikzpicture}
}
\end{subfigure}
\fi

\begin{subfigure}[b]{0.31\textwidth} %\centering
\resizebox{1\textwidth}{!}{
    \begin{tikzpicture}
      \begin{axis}
        [xmin = 300, xmax = 1000, xlabel={\# superpoints}, ylabel={$OOA$}, ymin = 93, ymax = 99, y label style={at={(+0.05,0.5)}}]
	
       \addplot [ptn] table [x = {N}, y = {ASA}] {\ptnvkitti};
        \addplot [slic] table [x = {N}, y = {ASA}] {\slicvkitti};
        \addplot [geof, mark=x] table [x = {N}, y = {ASA}] {\geofkitti};
        \addplot [cvpr] table [x = {N}, y = {ASA}] {\cvprkitti};
        \addplot [isprs] table [x = {N}, y = {ASA}] {\isprskitti};
      \end{axis}
    \end{tikzpicture}
    }
       \caption{OOA for vKITTI}
       \label{fig:OA:VKITTIRGB}
  \end{subfigure} 
  &
\begin{subfigure}[b]{0.31\textwidth} %\centering
\resizebox{1\textwidth}{!}{
    \begin{tikzpicture}
      \begin{axis}
        [xmin = 300, xmax = 1000, xlabel={\# superpoints}, ylabel={$BR$}, ymin = 60, ymax = 96, y label style={at={(+0.05,0.5)}}]
	 %\addplot [ptn] table [x = {N}, y = {BR}] {\ptnvkittid};
       \addplot [ptn] table [x = {N}, y = {BR}] {\ptnvkitti};
         \addplot [slic] table [x = {N}, y = {BR}] {\slicvkitti};
       \addplot [geof, mark=x] table [x = {N}, y = {BR}] {\geofkitti};
           \addplot [cvpr] table [x = {N}, y = {BR}] {\cvprkitti};
        \addplot [isprs] table [x = {N}, y = {BR}] {\isprskitti};
      \end{axis}
    \end{tikzpicture}
    }
       \caption{BR for vKITTI}
       \label{fig:BR:VKITTIRGB}
  \end{subfigure} 
  &
\begin{subfigure}[b]{.31\textwidth} %\centering
\resizebox{1\textwidth}{!}{
    \begin{tikzpicture}
      \begin{axis}
        [xmin = 300, xmax = 1000, xlabel={\# superpoints}, ylabel={$BP$} , ymin = 15, ymax = 65, y label style={at={(+0.05,0.5)}}]

        \addplot [ptn] table [x = {N}, y = {BP}] {\ptnvkitti};
         %\addplot [ptn] table [x = {N}, y = {BP}] {\ptnvkittid};
        \addplot [geof, mark=x] table [x = {N}, y = {BP}] {\geofkitti};
          \addplot [slic] table [x = {N}, y = {BP}] {\slicvkitti};
        \addplot [cvpr] table [x = {N}, y = {BP}] {\cvprkitti};
        \addplot [isprs] table [x = {N}, y = {BP}] {\isprskitti};
      \end{axis}
    \end{tikzpicture}
     }
     \caption{BP for vKITTI}
     \label{fig:BP:VKITTIRGB}
  \end{subfigure}
 
\end{tabular}
\end{center}

%% file: tables.tex
%auto-ignore
\setlength{\belowcaptionskip}{-8pt}
\begin{table}
\begin{tabular}{c|ccc}
\textbf{Method} & OA & mAcc & mIoU \\\hline
\multicolumn{4}{c}{6-fold cross validation}\\\hline
PointNet \cite{qi2017pointnet} in \cite{EngelmannKHL17_vkitty} & 78.5 & 66.2& 47.6 \\
Engelmann \etal in \cite{EngelmannKHL17_vkitty} & 81.1 & 66.4 & 49.7 \\
PointNet++ \cite{QiYSG17PointNetPP}  & 81.0 & 67.1& 54.5 \\
Engelmann \etal in \cite{engelmann2018} & 84.0 & 67.8 & 58.3 \\
SPG \cite{landrieu2017large} & 85.5 & 73.0 & 62.1 \\
PointCNN  \cite{li2018pointcnn} & \bf 88.1 & 75.6 & 65.4 \\
%PointSIFT \cite{jiang2018pointsift} & \bf 88.7 & - & \bf 70.2 \\
SSP + SPG (ours) &  87.9 & \bf 78.3 &  \bf 68.4\\
\hline
\multicolumn{4}{c}{Fold 5}\\
\hline
PointNet \cite{qi2017pointnet} in \cite{engelmann2018} & - & 49.0 & 41.1 \\
Engelmann \etal in \cite{engelmann2018} & 84.2 & 61.8 & 52.2 \\
pointCNN \cite{li2018pointcnn} & 85.9 & 63.9 & 57.3 \\
SPG \cite{landrieu2017large} & 86.4 & 66.5 & 58.0 \\
PCCN \cite{wang2018deep} & - & 67.0 & 58.3 \\
SSP + SPG (ours) & \bf 87.9 & \bf68.2 & \bf 61.7\\
\end{tabular}
\caption{Performance of different methods for the semantic segmentation task on the S3DIS dataset. The top table is for the 6-fold cross validation, the bottom table on the fifth fold only. %OA represents the overall accuracy, mAcc the mean accuracy per class, and mIoU the mean IoU.
}
\label{tab:s3dis:semseg}
\end{table}
\setlength{\abovecaptionskip}{-3pt}
\begin{table}\begin{center}
\begin{tabular}{c|ccc}
\bf Method& OA & mAcc & mIoU\\\hline
PointNet \cite{qi2017pointnet}
&79.7&47.0&34.4\\
Engelmann \etal in \cite{engelmann2018}
&79.7&57.6&35.6\\
Engelmann \etal in \cite{EngelmannKHL17_vkitty}
&80.6&49.7&36.2\\
3P-RNN \cite{YeLHDZ18_eccv}
&\bf87.8&54.1&41.6\\
SSP + SPG (ours) &84.3&\bf67.3&\bf52.0\\%\hline
%\multicolumn{4}{c}{xyz only}\\\hline
%%PointNet \cite{qi2017pointnet}
%&71.7&38.1&23.9\\
%Engelmann \etal in %\cite{EngelmannKHL17_vkitty}
%&73.9&46.7&29.8\\
%PointNet++\cite{QiYSG17PointNetPP}
%&77.0&40.0&29.9\\
%Engelmann \etal in %\cite{engelmann2018}
%&78.2&56.4&33.4\\
%3P-RNN \cite{YeLHDZ18_eccv}
%&79.6&49.2&34.5\\
%SSO + SPG (ours) &\bf 80.1&\bf %50.4&\bf 39.0%
\end{tabular}
\end{center}
\caption{Performance of different methods for the semantic segmentation task on the vKITTI dataset with 6-fold cross validation.}
\label{tab:vkitti:segsem}
\end{table}

%% file: supplementary.tex
%auto-ignore
\setlength{\belowcaptionskip}{5pt}
\setlength{\abovecaptionskip}{5pt}
\begin{table*}[h!]\centering
\begin{tabular}{ccc|cc}
\bf parameter & \bf shorthand & \bf section & \bf S3DIS & \bf vKITTI \\\hline
Local neighborhood size & $k$ & 3.1 & \multicolumn{2}{c}{20}\\
\# parameters & - &- & \multicolumn{2}{c}{13,816}\\
$\LCE$ configuration & - & 3.1 & \multicolumn{2}{c}{[32,128],[64,32,32,m]}\\
ST configuration & - & 3.1 & \multicolumn{2}{c}{[16,64],[32,16,4]}\\
Embeddings dimension & $m$ & 3.1 & \multicolumn{2}{c}{4}\\
Adjacency graph & $G$ & 3.2 & 5-nn & 5-nn + Delaunay\\
exponential edge factor & $\sigma$ & 3.2.1 & \multicolumn{2}{c}{0.5}  \\
intra-edge factor &$\tilde{\mu}$ & 3.2.3 & \multicolumn{2}{c}{5} \\
spatial influence&$\aspat$  & 3.4  & 0.2 & 0.02 \\
smallest superpoint & $n_\text{min}^{(1)}$ & 3.4 & 40 & 10 \\
epochs & - & - & \multicolumn{2}{c}{50}\\
decay event & - & - & 
\multicolumn{2}{c}{20,35,45}\\
\end{tabular}
\caption{Configuration of the embedding network for the S3DIS and vKITTI datasets.\\}
\label{tab:parameters}
\end{table*}
%--------------------------------
\section{Models configuration}
%--------------------------------
In this section, we give the full hyper-parameterization of all the networks used in the paper, for both oversegmentation and semantic segmentation tasks, and for both datasets.
%--------------------------------
\subsection{Models configuration for oversegmentation}
%--------------------------------
%
Our supervized oversegmentation model has a number of critical hyper-parameters to tune, given in \tabref{tab:parameters}. We detail here the rationale behind our choices.

\noindent
\textbf{Local neighborhood and adjacency graphs:} For both datasets, we find that setting the local neighborhood size to $20$ was enough for embeddings to successfully detect objects' border. Combined with our lightweight structure, this results in a very low memory load overall. The adjacency graph $G$ requires more attention depending on the dataset. For the dense scans of S3DIS, the $5$-nearest neighbors adjacency structure was enough to capture the connectivity of the input clouds. For the sparse scans of vKITTI, we added Delaunay edges \cite{delaunay1934sphere} (pruned at 50 cm) such that parallel scans lines would be connected.\\

\noindent
\textbf{Networks configuration:} For the $\LCE$ and the PointNet structure in the spatial transform, we find that shallow and wide architectures works better than deeper networks. We give in \tabref{tab:parameters} the size of the linear layers, before and after the maxpool operation. Over $250,000$ points can be embedded simultaneously on 11GB RAM in the training step, while keeping track of gradients.\\

\noindent
\textbf{Intra-edge factor:}
The graph-structured contrastive loss presented in  3.2.2 requires setting a weight $\mu$ determining the influence of inter-edges with respect to intra-edge. Since most edges of $G$ are intra-edges in practice, we define $\tilde{\mu}$ such that $\mu=\tilde{\mu} c$ with $c={\mid E \mid}/{\mid V \mid}$ the average connectivity of $G$. Note that $c$ can be determined directly from the construction of the adjacency graph (it is equal to $k$ in a $k$-nearest neighbor graph for example). A value of $\tilde{\mu}=1$ means that the total influence in $\ell$ of inter-edges and intra-edges are identical. Since we are interested in oversegmentation, we set $\tilde{\mu}$ to $5$ in all our experiments, but note that the network is not very sensitive to this parameter, as demonstrated experimentally: a value of $\tilde{\mu}=3$ gives a relative performance of $(-0.2,-0.6,+1.5)$ while a value of $8$ gives $(+0.1,-0.5,+1.4)$.\\

\noindent
\textbf{Regularization Strength:} The generalized minimal partition problem defined in 3.2.1 requires setting the regularization strength factor $\lambda$, determining the cost of  edges crossing superpoints. We remark that the $\LCE$ produces embeddings of points with an euclidean distance of at least $1$ over predicted objects' borders. Some calculus shows us that for a $\lambda\leq 1/(2 c)$, the solution $f^\star$ of (8) should predict superpoints borders at all edges whose vertices have a difference of embeddings of at least $1$ (note that there is no guarantee that the greedy $\ell_0$-cut pursuit algorithm will indeed predict a border). We use this value to define a normalized regularization strength $\tilde{\lambda}$ such that $ \lambda = \tilde{\lambda}/(4c)$, whose default value is $1$.\\

\begin{table*}[h!]\centering
\begin{tabular}{c|cc}
\bf parameter  &  \bf S3DIS & \bf vKITTI \\\hline
\# parameters & 278,897
 & 118,737
\\
Superpoint embedders configuration &[[64,64,128,128,256], [256,64,32]] & [[64,64,128,256], [128,32,32]]\\
STN configuration &[[64,64,128], [128,64]] & [[32,32,64], [64,32]\\
subsampling hops & \multicolumn{2}{c}{4}\\
max SPgraph size & \multicolumn{2}{c}{768}\\
$\tilde{\lambda}$ & 0.1 & 0.5 \\
$n_\text{min}$ & 25 & 15 \\
epochs & 350 & 100 \\
decay event & 180,250,280,320 & 40,50,60,70,80\\
\end{tabular}
\caption{Configuration of the semantic segmentation network. All values not mentioned in this table use default parameters from \cite{landrieu2017large}}
\label{tab:parameters_semseg}
\end{table*}

\noindent
\textbf{Regularization path:} To obtain the regularization paths in Figure 7, we first train the network with a regularization strength of $\tilde{\lambda}=1$ (see 3.2.2). We then compute partitions with $\tilde{\lambda}$ varying from $0.2$ to $6$ with no fine-tuning required.\\

\noindent
\textbf{Smallest superpoint:} 
To automatically select a minimal superpoint size (in number of points) appropriate to the coarseness of the segmentation, we heuristically set:
$$
n_\text{min}^{\tilde{\lambda}}=\Bra{(\max\Pa{\frac12 n_\text{min}^{(1)},n_\text{min}^{(1)} + \frac12 n_\text{min}^{(1)}\log(\tilde{\lambda})}}
$$
where $n_\text{min}^{(1)}$ is a dataset-specific minimum superpoints size for $\tilde{\lambda}=1$. For example, for $n_\text{min}^{(1)}=50$, the smallest superpoint allowed for a small regularization strength $\tilde{\lambda}= 0.2$ will be $33$, while it is $70$ for the coarse partition obtained with $\tilde{\lambda}=6$. While specific applications may require setting up this variable manually, this allowed us to produce the regularization paths in Figure 7 while only varying $\tilde{\lambda}$.\\

\noindent
\textbf{Optimization:}
Given the small size of our network, we train our network for a short number of epochs (see \tabref{tab:parameters}), with decay events set at $0.7$. We use Adam optimizer \cite{kingma2014adam} with gradient clipping at $1$ \cite{goodfellow2016deep}. Training takes around 2 hours per fold on our $11$GB VRAM $1080$Ti GPU.\\

\noindent
\textbf{Mini-batches:}
For graph-based clustering, the training phase processes batches of $16$ point clouds at once, for which a subgraph of size $10\,000$ points is extracted. For the clustering-based segmentation, which is more memory intensive, and since subgraphs have to be larger to be meaningfully covered by the initial voxels, we set a batch size of $1$ and a subgraph of $100\,000$. As a consequence, we replace the batchnorm layers of the $\LCE$s by group norms with $4$ groups \cite{wu2018group}.\\

\noindent
\textbf{Augmentation:} In order to build more robust networks, we added Gaussian noise of deviation $0.03$ clamped at $0.1$ on the normalized position and color of neighborhood clouds. We also added random rotation of the input clouds for the network to learn rotation invariance. To preserve orientation information, the clouds are rotated as a whole instead of each neighborhood. This allows the spatial transform to detect change in orientation, which can be used to detect borders.
%**********************************
\begin{table*}[ht!]\begin{center}
\resizebox{1\linewidth}{!}{
\begin{tabular}{|c|C{0.04\textwidth}C{0.04\textwidth}C{0.04\textwidth}|C{0.04\textwidth}C{0.04\textwidth}C{0.04\textwidth}C{0.04\textwidth}C{0.05\textwidth}C{0.05\textwidth}C{0.04\textwidth}C{0.04\textwidth}C{0.04\textwidth}C{0.06\textwidth}C{0.04\textwidth}C{0.04\textwidth}C{0.04\textwidth}|}\hline
\bf Method & OA & mAcc & mIoU & ceiling & floor & wall & beam & column & window & door & chair & table & bookcase & sofa & board & clutter\\\hline
A5 PointNet \cite{qi2017pointnet}&--&49.0&41.1&88.8&97.3&69.8&\bf 0.1 &3.9&46.3&10.8&52.6&58.9&40.3&5.9&26.4&33.2\\ 
A5 SEGCloud \cite{tchapmi2017segcloud}&--&57.4&48.9&90.1 &96.1 &69.9&0.0&18.4&38.4&23.1&75.9&70.4&58.4&40.9 &13.0&41.6\\
A5 PointCNN \cite{li2018pointcnn} & 85.9 &63.9& 57.3& \bf 92.3& \bf 98.2 &79.4& 0.0 &17.6& 22.8& 62.1& 80.6 & 74.4 & 66.7 & 31.7 &\bf 62.2& \bf 56.7 \\
A5 SPG \cite{landrieu2017large}
&86.4&66.5&58.0&
89.4&96.9&78.1&0.0&\bf 42.8
&48.9&\bf61.6&84.7& 75.4&
69.8&\bf52.6&2.1&52.2\\
A5 SSP + SPG (ours) & \bf 87.9 & \bf 68.2 & \bf 61.7 & 91.9 & 96.7 & \bf 80.8 & \bf 0.0 & 28.8 & \bf 60.3 &  57.2 & \bf 85.5 & \bf76.4 & \bf 70.5 & 49.1 & 51.6 & 53.3 \\
\hline
PointNet \cite{qi2017pointnet} in \cite{Engelmann17_3dsemseg}&78.5&66.2&47.6&88.0&88.7&69.3&42.4&23.1&47.5&51.6&42.0&54.1&38.2&9.6&29.4&35.2\\
Engelmann \etal~\cite{Engelmann17_3dsemseg}&81.1&66.4&49.7&90.3&92.1&67.9&44.7&24.2&52.3&51.2&47.4&58.1&39.0&6.9&30.0&41.9\\
Engelamnn in \cite{engelmann2018know}& 84.0 &67.8& 58.3 &92.1 &90.4 &78.5& 37.8& 35.7& 51.2&65.4 & 61.6 &64.0 &51.6&25.6  & 49.9&53.7 \\
SPG \cite{landrieu2017large}&85.5&73.0&62.1&89.9&95.1&76.4& 62.8& 47.1& 55.3& 68.4& 73.5& 69.2& 63.2& 45.9&8.7& 52.9\\
PointCNN \cite{li2018pointcnn}&\bf 88.1& 75.6 &65.4 &\bf 94.8 &\bf 97.3& 75.8& \bf 63.3& 51.7 & 58.4&57.2&69.1 & \bf 71.6 & 61.2 & 39.1& 52.2& 58.6\\
SSP + SPG (ours) & 87.9 & \bf 78.3 & \bf 68.4 & 91.7 & 95.5 & \bf 80.8 &  62.2 & \bf 54.9 & \bf 58.8 & \bf 68.4 & \bf 78.4 & 69.2  & \bf 64.3 & \bf 52.0 & \bf 54.2 & \bf59.2 \\\hline
\end{tabular}}
\end{center}
\caption{Results on the S3DIS dataset on fold ``Area 5'' (top) and micro-averaged over all 6 folds (bottom). Intersection over union is shown split per class, with the highest value over all methods in bold.}
\label{tab:results_S3DIS}
\end{table*}
%--------------------------------
\subsection{Models configuration for semantic segmentation}
%--------------------------------
We used the open-source superpoint-graph implementation \url{github/loicland/superpoint-graph} without any modification beyond changing the oversegmentation step and some changes in the hyper-parameters. The full parameterization is given in \tabref{tab:parameters_semseg}.

To compensate for the edges missed by the $\ell_0$-cut pursuit approximation, due in part to its ignoring the spherical nature of the embeddings, we set the regularization strength $\tilde{\lambda}$ lower than 1 for both datasets. This help improve the accuracy and border recall. The subsequent decrease in border precision is compensated by the fact that the SPG, through its context leveraging module, can learn to propagate the semantic information to small superpoints. For the same reason, we chose a lower superpoint size for S3DIS from the segmentation experiments.

We extended the superpoint graph subsampling threshold to $4$-hops instead of $3$, because our method SSP tends to produce thin components near interfaces. Since the vKITTI dataset is much smaller than S3DIS, we chose smaller networks to mitigate overfitting.
%

%--------------------------------
\section{Residual Point Embedder}
%--------------------------------
We have tested an alternative configuration for the local point embedded, in which they were stacked in layers, similarly to the classical convolutional architecture for images. We first introduce a slightly changed architecture, the Residual Point Embedder $\RPE$, whose design is based on an $\LCE$ but takes a supplementary input $e_\text{ini}$. Instead of computing a new embedding, the $\RPE$ computes a residual \eqref{eq:res2} which is added to this initial embedding before normalization \eqref{eq:res_norm}:
\begin{align}\label{eq:res2}
    &R(X_i,x_i) = \MLP_2\Pa{[\max\Pa{\MLP_1(X_i)}, x_i]}\\\label{eq:res_norm}
    &\RPE(x_i,X_i,e_\text{ini})\ = \Ltwo \Pa{e_\text{ini}+R(X_i,x_i)}
\end{align}
The second change is the layers architecture. The $\RPE$s in the first layer compute the embeddings from the local geometric and radiometric information alone, and their initial embedding is set to $0$ (\ref{eq:emb0}) (such that they behave exactly like $\LCE$s). The $\RPE$s in subsequent layers compute new embeddings from the local radiometry and geometry as well as the embeddings computed at the previous layer of the points neighbors $E_i^{t}$ (\ref{eq:embt}). Note that for a point to be processed by a layer, all its neighbors must have been embedded by the previous layer. This allows the $\RPE$s to have increasingly broader receptive fields, and to correct errors that might have been done by previous layers. Note that the geometric information are only processed by the spatial transform once, cascading its values to all residual layers.
\begin{align}\label{eq:emb0}
 &\upp{e_i}{0}=\upp{\RPE}{0}([\Pit,R_i],[\pit,r_i],0)\\\label{eq:embt}
 &\upp{e_i}{t+1}=\upp{\RPE}{t}([\Pit,\upp{E_i}{t}],[\pit,r_i,\upp{e_i}{t}],\upp{e_i}{t})
 \end{align}
Alternatively, all initial embeddings can be set to $0$, which means that each layer computes a new embedding from the local position and the embeddings of the previous layers. As mentioned in the ablation study, while these networks did perform well, their benefits shrink when a simple $\LCE$ is given as many parameters.
%--------------------------------
\section{Detailed results and illustration}
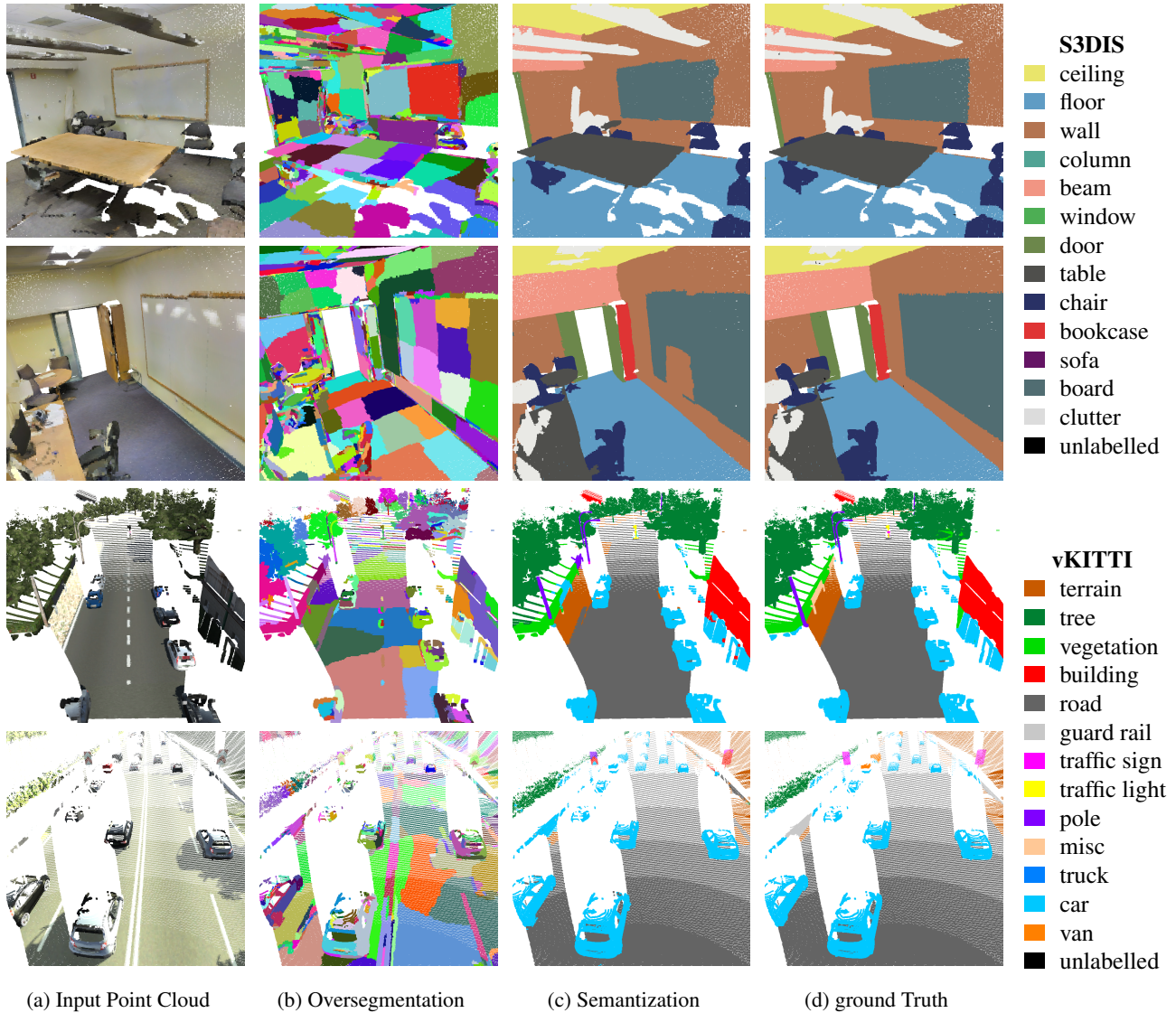
\begin{figure*}
\input{illustration_semseg}
\caption{Illustration of the results on the semantic segmentation. In the first row we show a successful semantization for a complex scene of S3DIS. In the second row, we show a failure case in which a white board is oversegmented in too many small superpoints. This makes their classification harder by the semantic segmentation network.
In the third row we see a successful semantization of an urban outdoor scene from vKITTI. On the fourth row, we can observe in the background road signs with high color contrasts, which are segmented in small superpoints. This makes them very hard to classify and they are missed by the semantic segmentation algorithm.}
\label{fig:semseg}
\end{figure*}

%--------------------------------
We present in \tabref{tab:results_S3DIS} the per-class IoU for the S3DIS dataset. We illustrate the semantic segmentation results in \figref{fig:semseg}.
We also made a video illustration which can be accessed at \url{https://youtu.be/bKxU03tjLJ4}.
%====================================

%% file: illustration_semseg.tex
%auto-ignore
\begin{tabular}[b]{ccccc}
\begin{subfigure}[b]{0.2\textwidth}
\parbox[b]{1\textwidth}{
    \begin{tabular}[b]{c}
    \includegraphics[width=1\textwidth]{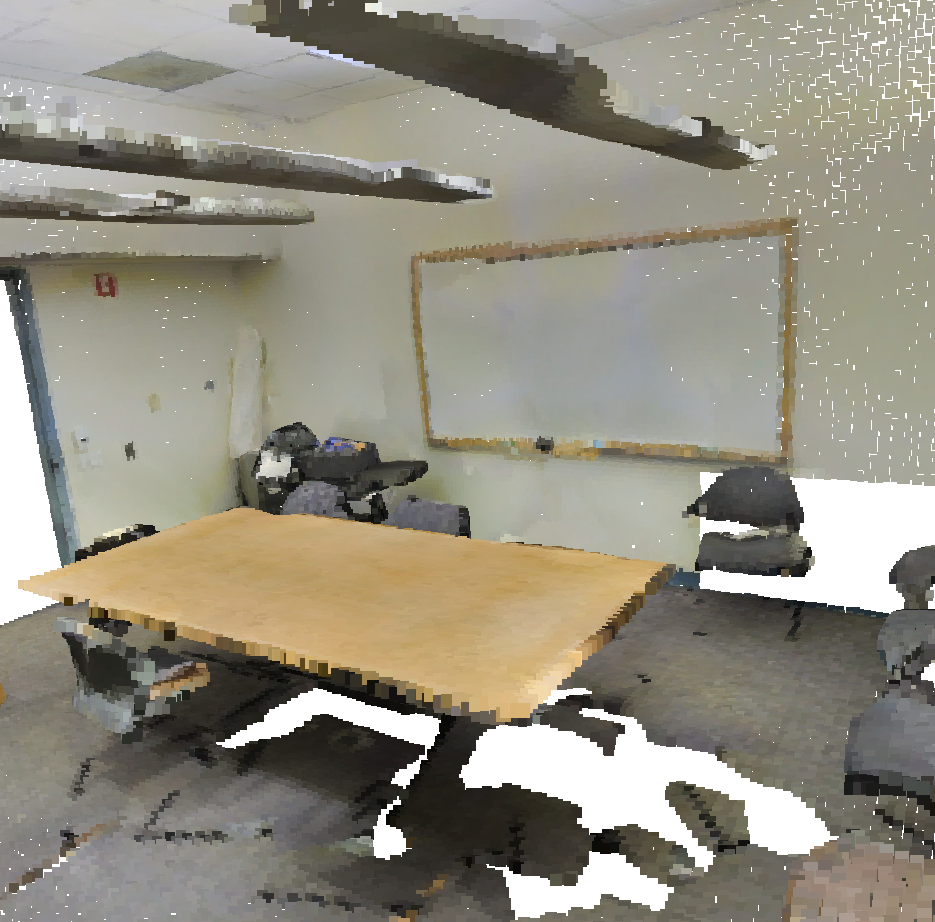}\\
    \includegraphics[width=1\textwidth]{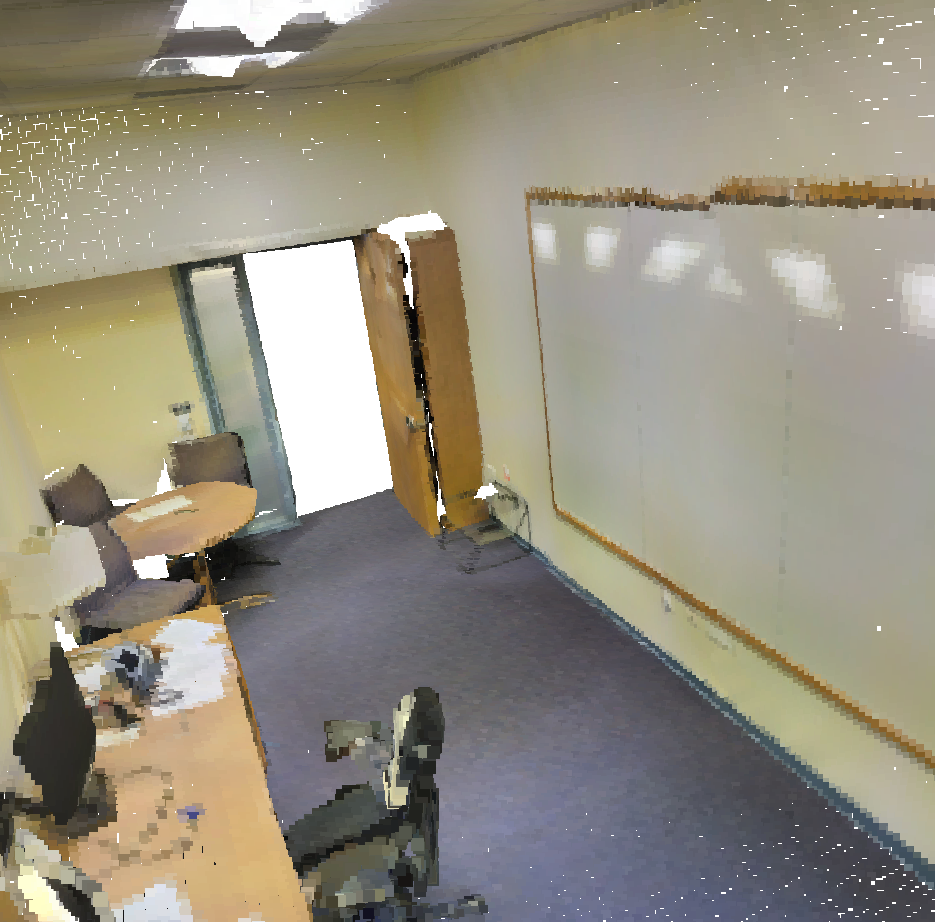}\\
    \includegraphics[width=1\textwidth]{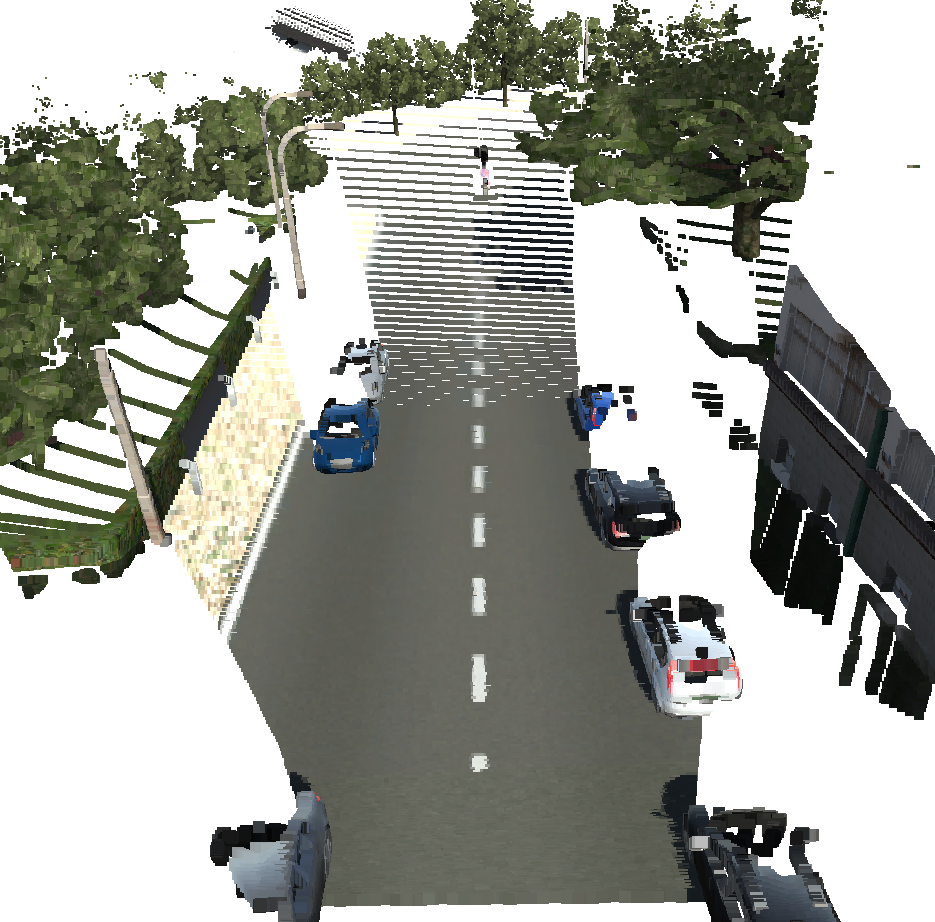}\\
    \includegraphics[width=1\textwidth]{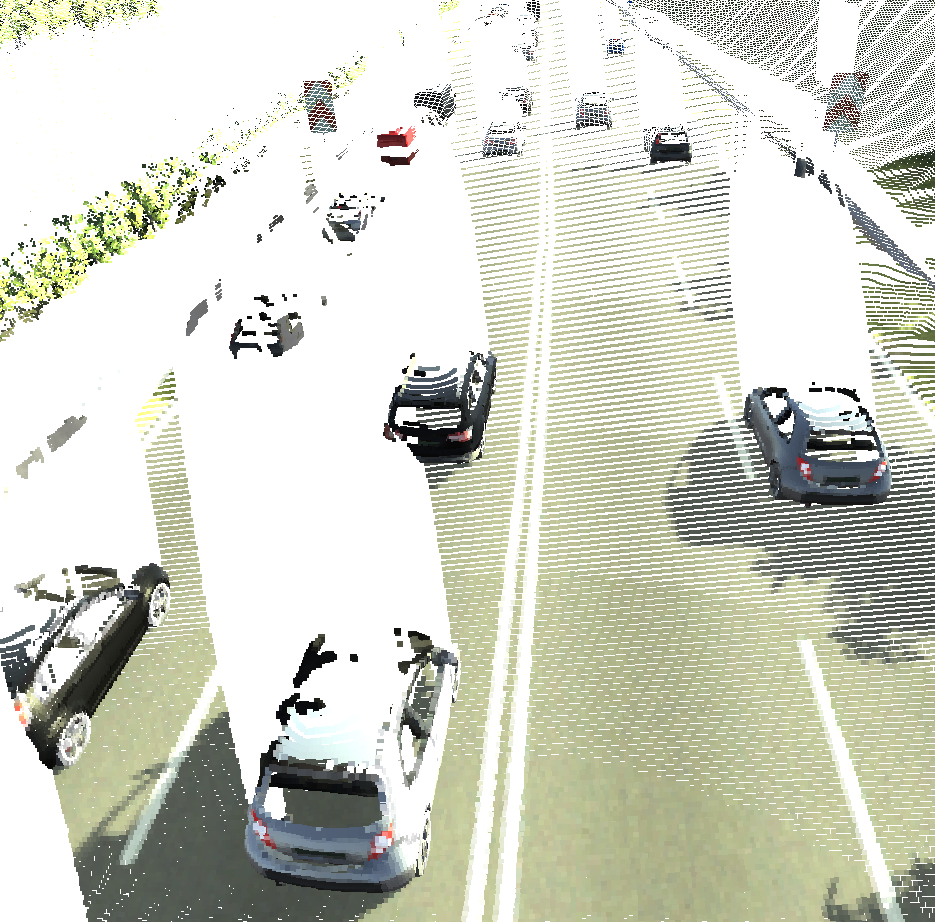}
    \end{tabular}
    \caption{Input Point Cloud}
    \label{fig:semseg:rgb}
    }\end{subfigure}
&
\begin{subfigure}[b]{0.2\textwidth}
\parbox[b]{1\textwidth}{
\begin{tabular}[b]{c}
\includegraphics[width=1\textwidth]{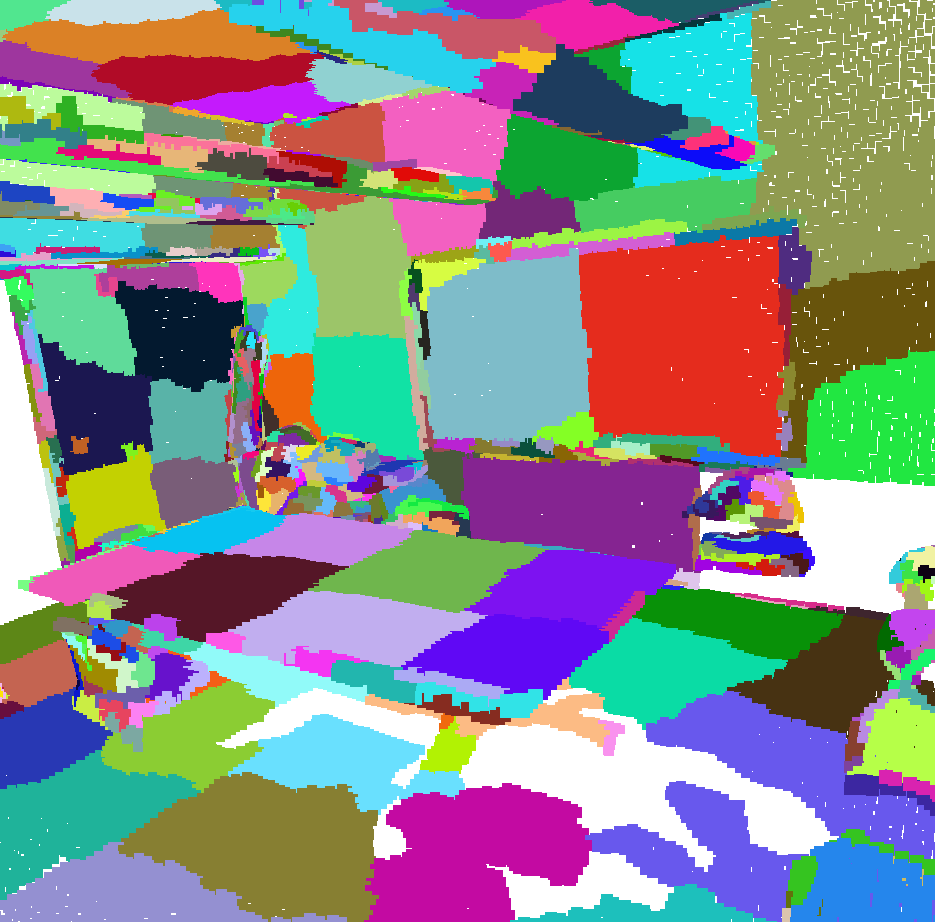}\\
\includegraphics[width=1\textwidth]{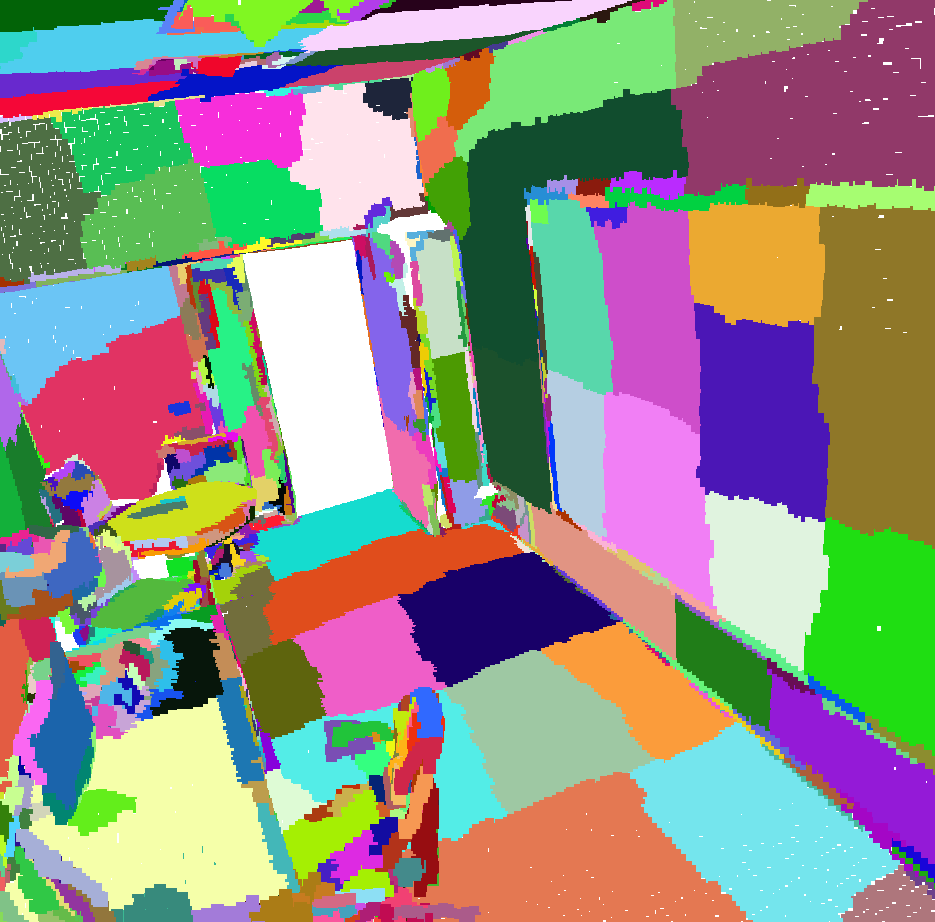}\\
\includegraphics[width=1\textwidth]{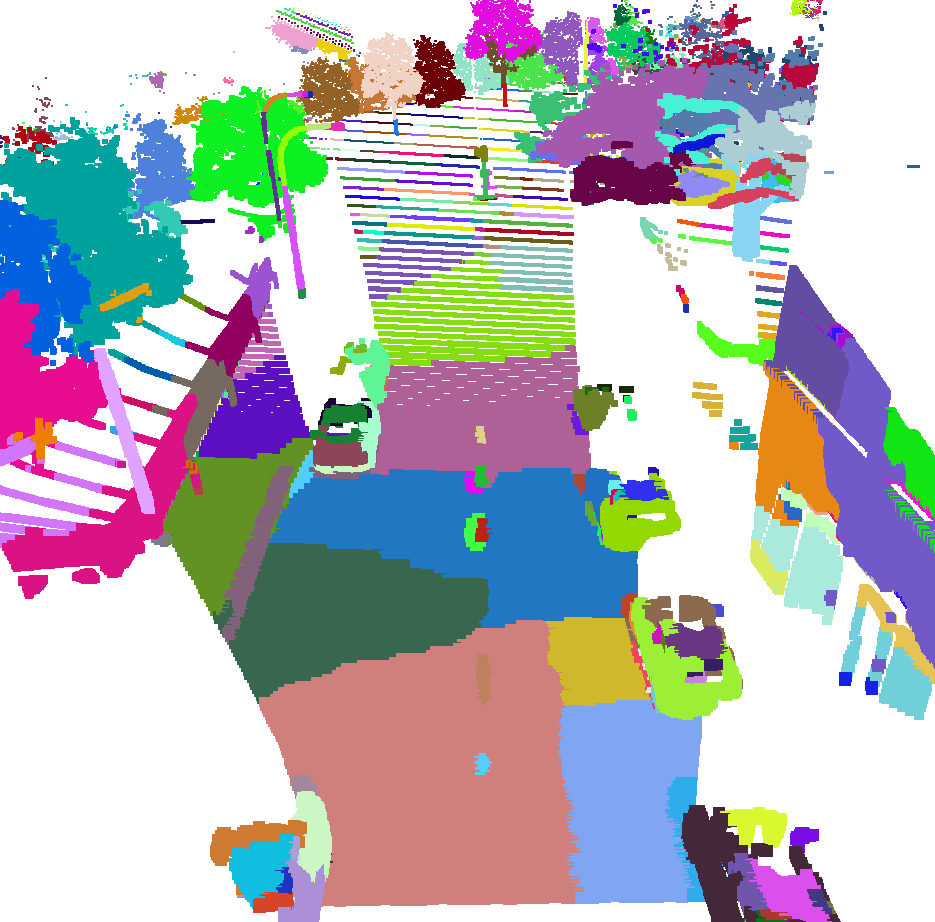}\\
\includegraphics[width=1\textwidth]{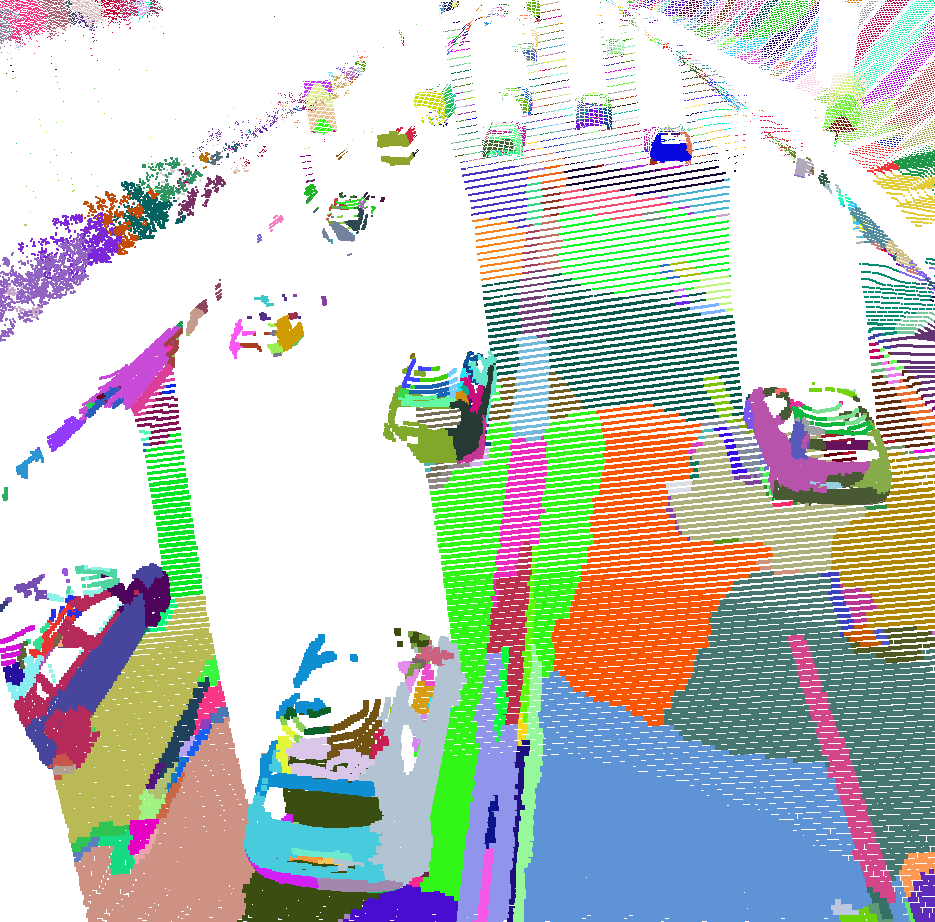}
\end{tabular}
   \caption{Oversegmentation}
    \label{fig:semseg:par}
 }\end{subfigure}
&
\begin{subfigure}[b]{0.2\textwidth}
\parbox[b]{1\textwidth}{
\begin{tabular}[b]{c}
\includegraphics[width=1\textwidth]{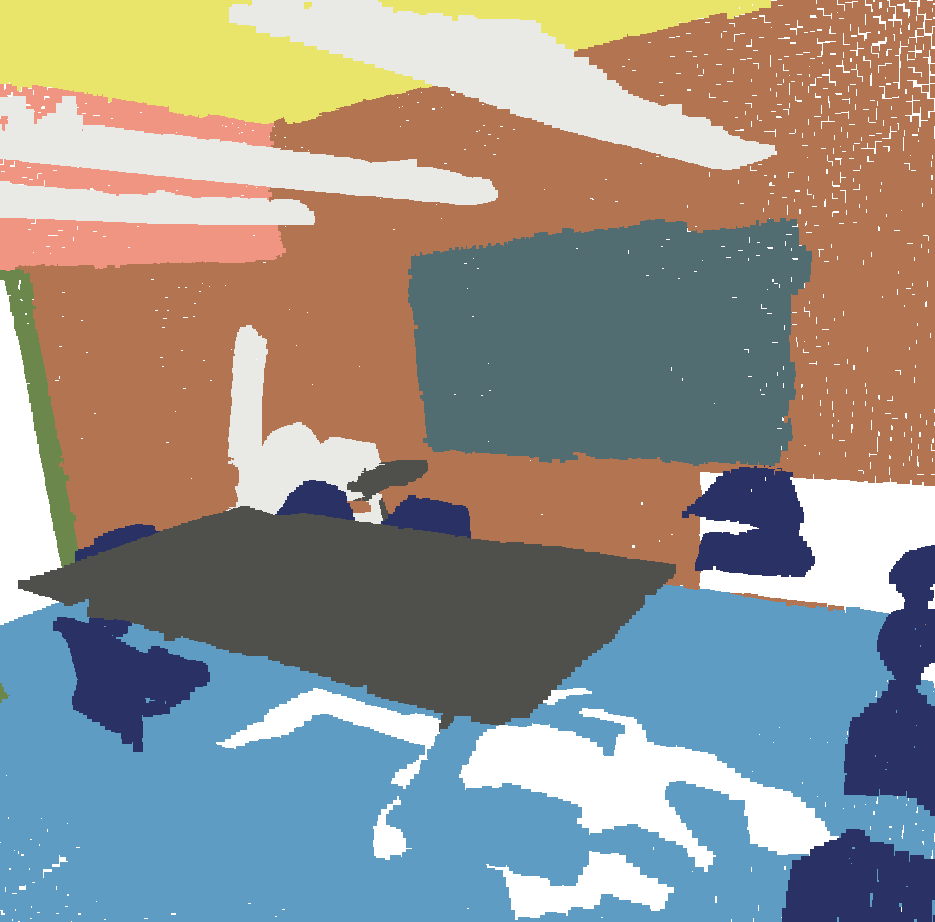}\\
\includegraphics[width=1\textwidth]{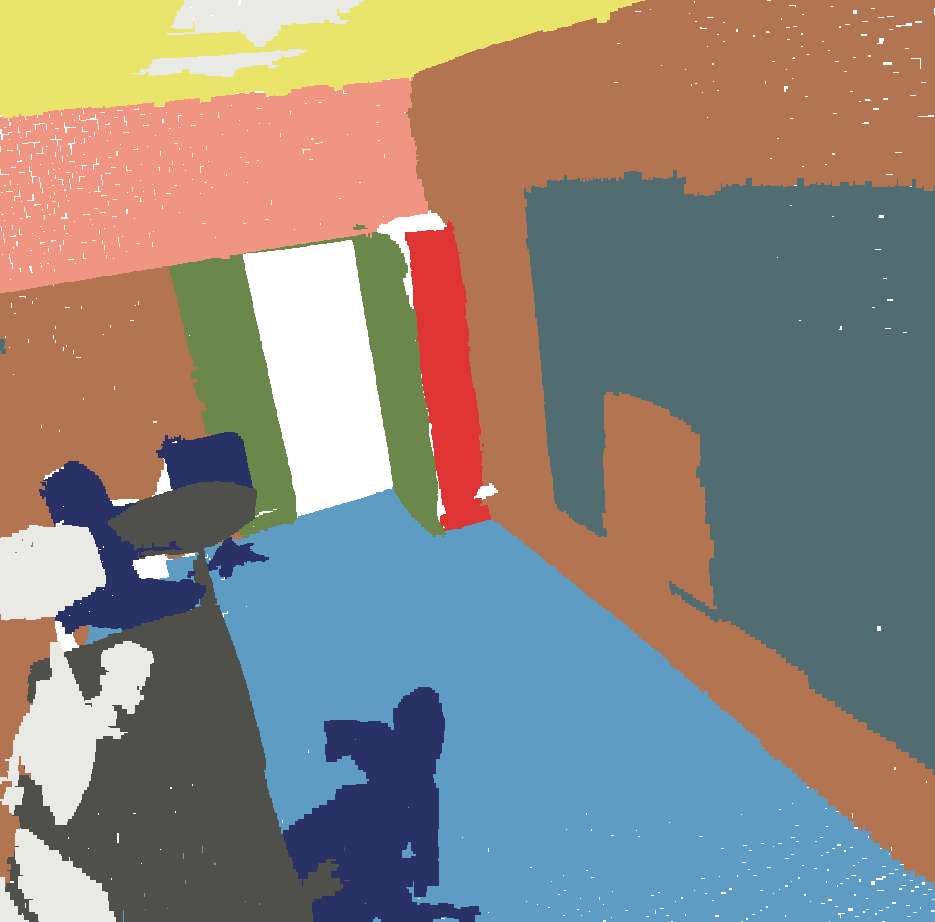}\\
\includegraphics[width=1\textwidth]{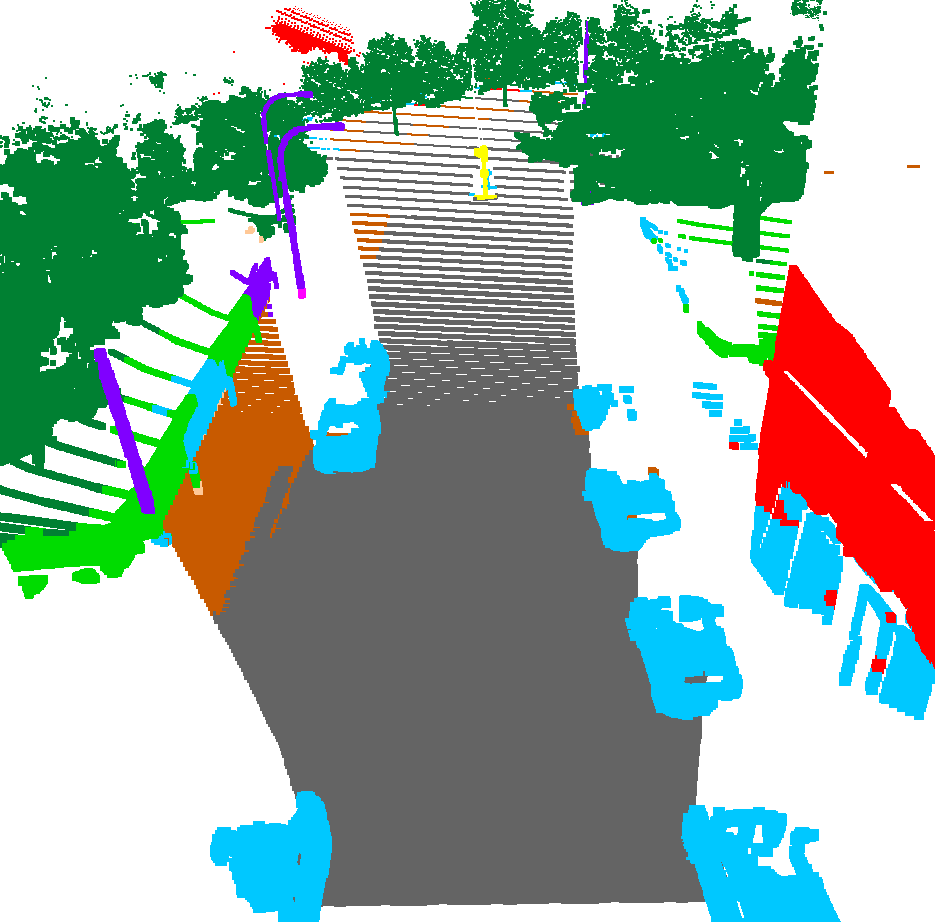}\\
\includegraphics[width=1\textwidth]{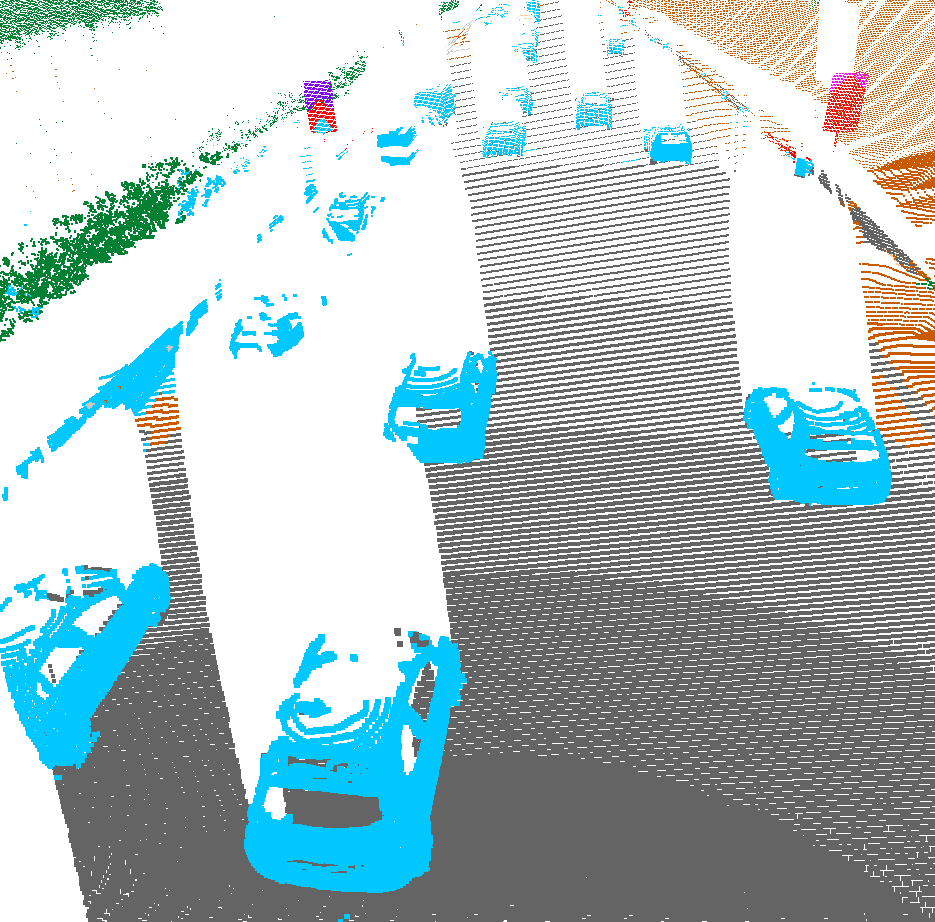}
\end{tabular}
   \caption{Semantization}
    \label{fig:semseg:pred}
 }\end{subfigure}
&
\begin{subfigure}[b]{0.2\textwidth}
\parbox[b]{1\textwidth}{
\begin{tabular}[b]{c}
\includegraphics[width=1\textwidth]{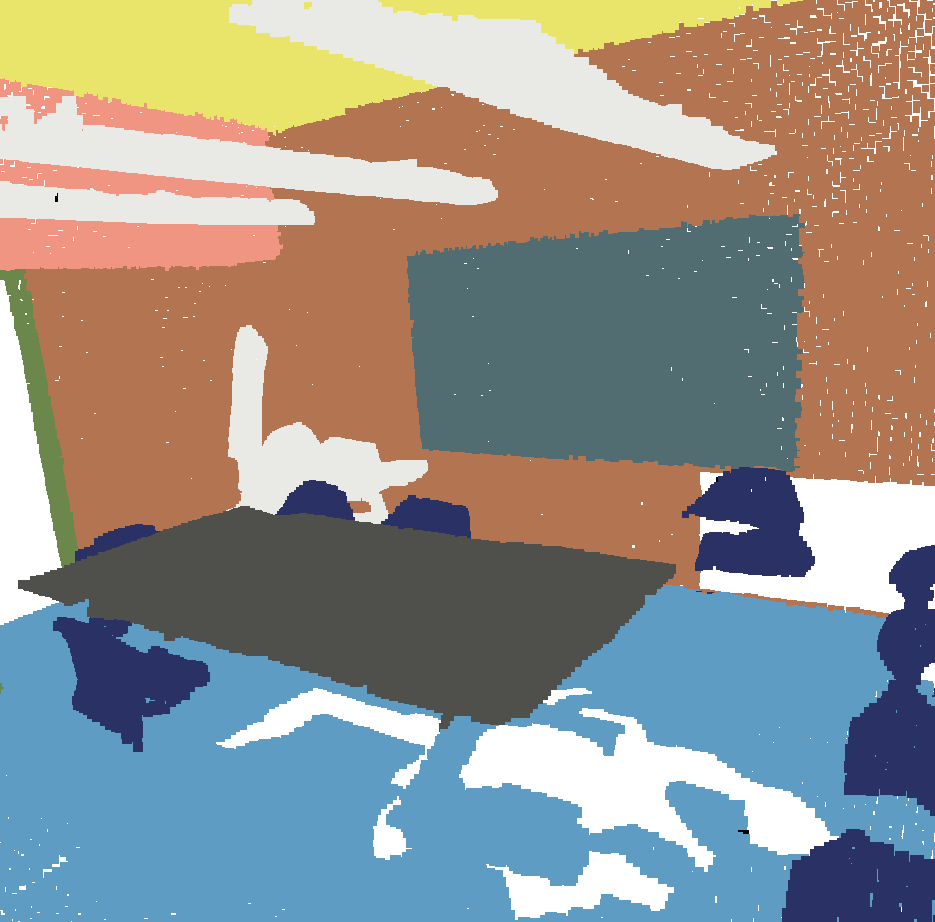}\\
\includegraphics[width=1\textwidth]{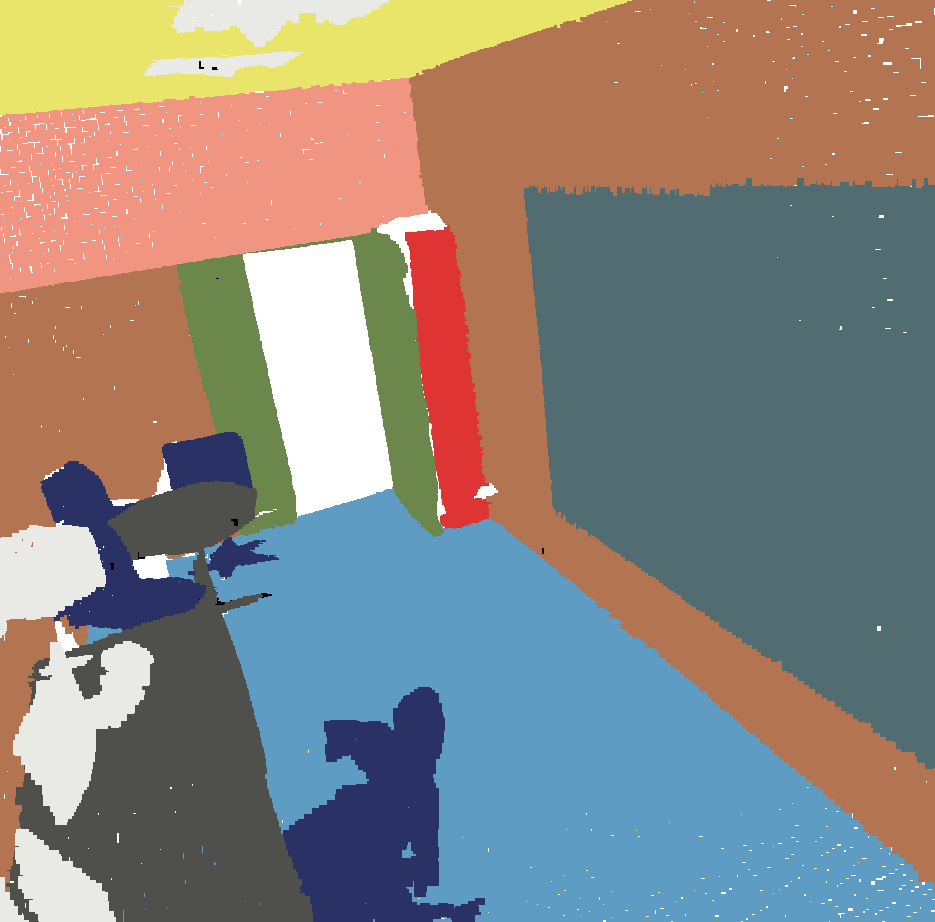}\\
\includegraphics[width=1\textwidth]{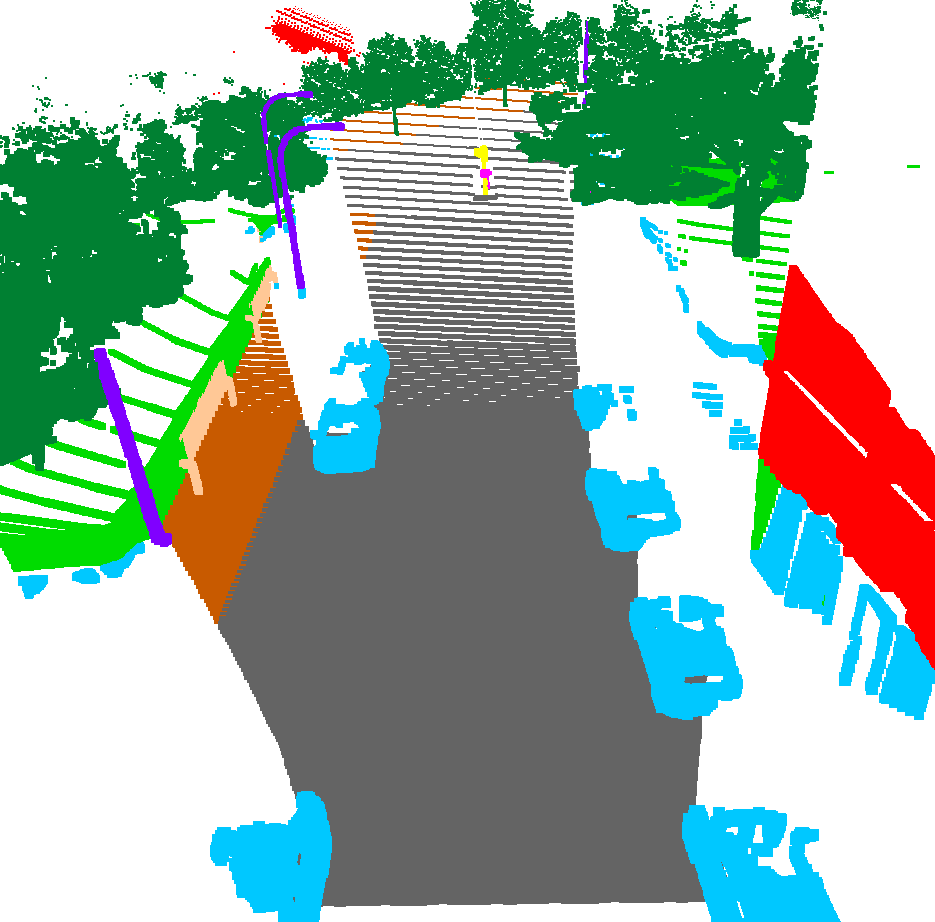}\\
\includegraphics[width=1\textwidth]{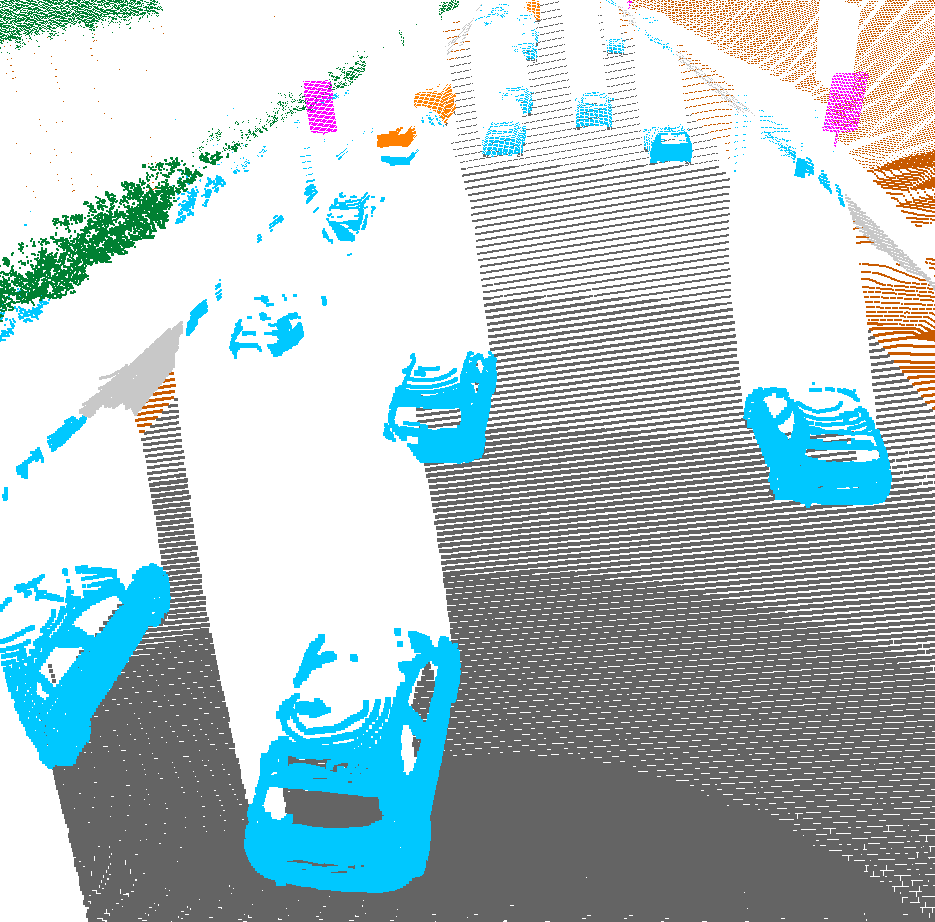}
\end{tabular}
   \caption{ground Truth}
    \label{fig:semseg:gt}
 }\end{subfigure}
 &
 \begin{subfigure}[b]{0.2\textwidth}
 \parbox[b]{1\textwidth}{
 \begin{tabular}[b]{rl}
 \multicolumn{2}{c}{\textbf{S3DIS}} \\
 \definecolor{tempcolor9}{RGB}{233,229,107}
 \tikz\node at (0,0) [rectangle, minimum width = 3mm, draw = none, fill = tempcolor9] (n){};
 & ceiling\\
 \definecolor{tempcolor10}{RGB}{95,156,196}
 \tikz\node at (0,0) [rectangle, minimum width = 3mm, draw = none, fill = tempcolor10] (n){};
 & floor\\
 \definecolor{tempcolor11}{RGB}{179,116,81}
 \tikz\node at (0,0) [rectangle, minimum width = 3mm, draw = none, fill = tempcolor11] (n){};
 & wall\\
 \definecolor{tempcolor12}{RGB}{81,163,148}
 \tikz\node at (0,0) [rectangle, minimum width = 3mm, draw = none, fill = tempcolor12] (n){};
 & column\\
 \definecolor{tempcolor13}{RGB}{241,149,131}
 \tikz\node at (0,0) [rectangle, minimum width = 3mm, draw = none, fill = tempcolor13] (n){};
 & beam\\
 \definecolor{tempcolor14}{RGB}{77,174,84}
 \tikz\node at (0,0) [rectangle, minimum width = 3mm, draw = none, fill = tempcolor14] (n){};
 & window\\
 \definecolor{tempcolor15}{RGB}{108,135,75}
 \tikz\node at (0,0) [rectangle, minimum width = 3mm, draw = none, fill = tempcolor15] (n){};
 & door\\
 \definecolor{tempcolor16}{RGB}{79,79,76}
 \tikz\node at (0,0) [rectangle, minimum width = 3mm, draw = none, fill = tempcolor16] (n){};
 & table\\
 \definecolor{tempcolor17}{RGB}{41,49,101}
 \tikz\node at (0,0) [rectangle, minimum width = 3mm, draw = none, fill = tempcolor17] (n){};
 & chair\\
 \definecolor{tempcolor18}{RGB}{223,52,52}
 \tikz\node at (0,0) [rectangle, minimum width = 3mm, draw = none, fill = tempcolor18] (n){};
 & bookcase\\
 \definecolor{tempcolor19}{RGB}{100,20,100}
 \tikz\node at (0,0) [rectangle, minimum width = 3mm, draw = none, fill = tempcolor19] (n){};
& sofa\\
 \definecolor{tempcolor20}{RGB}{81,109,114}
 \tikz\node at (0,0) [rectangle, minimum width = 3mm, draw = none, fill = tempcolor20] (n){};
 & board\\
 \definecolor{tempcolor21}{RGB}{220,220,220}
  \tikz\node at (0,0) [rectangle, minimum width = 3mm, draw = none, fill = tempcolor21] (n){};
 & clutter\\ 
 \definecolor{tempcolor22}{RGB}{0,0,0}
\tikz\node at (0,0) [rectangle, minimum width = 3mm, draw = none, fill = tempcolor22] (n){};
 & unlabelled \\
 ~\\~\\~\\
 \multicolumn{2}{c}{\textbf{vKITTI}} \\
 \definecolor{tempcolor23}{RGB}{200,90,0}
 \tikz\node at (0,0) [rectangle, minimum width = 3mm, draw = none, fill = tempcolor23] (n){};
 & terrain\\
 \definecolor{tempcolor24}{RGB}{0,128,50}
 \tikz\node at (0,0) [rectangle, minimum width = 3mm, draw = none, fill = tempcolor24] (n){};
 & tree\\
 \definecolor{tempcolor25}{RGB}{0,220,0}
 \tikz\node at (0,0) [rectangle, minimum width = 3mm, draw = none, fill = tempcolor25] (n){};
 & vegetation\\
 \definecolor{tempcolor26}{RGB}{255,0,0}
 \tikz\node at (0,0) [rectangle, minimum width = 3mm, draw = none, fill = tempcolor26] (n){};
 & building\\
 \definecolor{tempcolor27}{RGB}{100,100,100}
 \tikz\node at (0,0) [rectangle, minimum width = 3mm, draw = none, fill = tempcolor27] (n){};
 & road\\
 \definecolor{tempcolor28}{RGB}{200,200,200}
 \tikz\node at (0,0) [rectangle, minimum width = 3mm, draw = none, fill = tempcolor28] (n){};
 & guard rail\\
 \definecolor{tempcolor29}{RGB}{255,0,255}
 \tikz\node at (0,0) [rectangle, minimum width = 3mm, draw = none, fill = tempcolor29] (n){};
 & traffic sign\\
 \definecolor{tempcolor30}{RGB}{255,255,0}
 \tikz\node at (0,0) [rectangle, minimum width = 3mm, draw = none, fill = tempcolor30] (n){};
 & traffic light\\
 \definecolor{tempcolor31}{RGB}{128,0,255}
 \tikz\node at (0,0) [rectangle, minimum width = 3mm, draw = none, fill = tempcolor31] (n){};
 & pole\\
 \definecolor{tempcolor32}{RGB}{255,200,150}
 \tikz\node at (0,0) [rectangle, minimum width = 3mm, draw = none, fill = tempcolor32] (n){};
 & misc\\
 \definecolor{tempcolor33}{RGB}{0,128,255}
 \tikz\node at (0,0) [rectangle, minimum width = 3mm, draw = none, fill = tempcolor33] (n){};
 & truck\\
 \definecolor{tempcolor34}{RGB}{0,200,255}
 \tikz\node at (0,0) [rectangle, minimum width = 3mm, draw = none, fill = tempcolor34] (n){};
 & car\\
 \definecolor{tempcolor35}{RGB}{255,128,0}
 \tikz\node at (0,0) [rectangle, minimum width = 3mm, draw = none, fill = tempcolor35] (n){};
 & van\\
 \definecolor{tempcolor36}{RGB}{0,0,0}
 \tikz\node at (0,0) [rectangle, minimum width = 3mm, draw = none, fill = tempcolor36] (n){};
 & unlabelled\\~\\~\\
 \end{tabular}
  \phantomcaption
 }
\end{subfigure}
\end{tabular}